\documentclass{egpubl}
 
 \CGFStandardLicense
\usepackage[T1]{fontenc}
\usepackage{dfadobe}  
\usepackage{cite}  
\BibtexOrBiblatex
\electronicVersion
\PrintedOrElectronic
\ifpdf \usepackage[pdftex]{graphicx} \pdfcompresslevel=9
\else \usepackage[dvips]{graphicx} \fi

\usepackage{egweblnk} 

\usepackage{microtype}
\usepackage{multicol, multirow}
\usepackage{amsfonts}
\usepackage{amsmath}
\usepackage{array}
\usepackage{wrapfig} 
 \usepackage{booktabs}
\usepackage[capitalize]{cleveref}
\crefname{paragraph}{Sec.}{Secs.}
\crefname{section}{Sec.}{Secs.}
\Crefname{section}{Section}{Sections}
\Crefname{table}{Table}{Tables}
\crefname{table}{Tab.}{Tabs.}
\Crefname{figure}{Figure}{Figures}
\crefname{figure}{Fig.}{Figs.}
\Crefname{chapter}{Chapter}{Chapters}
\crefname{chapter}{Ch.}{Chs.}

\providecommand{\eg}[0]{\emph{e.g.}, }

\providecommand{\ie}[0]{\emph{i.e.}, }

\providecommand{\etal}[0]{\emph{et~al.}}

\usepackage{xcolor}
\usepackage{colortbl}
\definecolor{1st}{RGB}{102,194,164} 
\definecolor{2nd}{RGB}{178,226,226} 
\definecolor{3rd}{RGB}{237,248,251} 
\definecolor{1stText}{RGB}{57,146,116}

\title[D-NPC]%
      {D-NPC: Dynamic Neural Point Clouds for \\ Non-Rigid View Synthesis from Monocular Video}

\author[M. Kappel et al.]
{\parbox{\textwidth}{\centering Moritz Kappel$^{1}$\orcid{0000-0001-9507-5141},
 Florian Hahlbohm$^{1}$\orcid{0009-0004-8710-1433}, 
 Timon Scholz$^{1}$\orcid{0009-0004-3635-190X}, 
 Susana Castillo$^{1}$\orcid{0000-0003-1245-4758},\\
 Christian Theobalt$^{2}$\orcid{0000-0001-6104-6625},
 Martin Eisemann$^{1}$\orcid{0000-0002-8673-4405}, 
 Vladislav Golyanik$^{2}$\orcid{0000-0003-1630-2006}, 
 and Marcus Magnor$^{1}$\orcid{0000-0003-0579-480X}
        }
        \\
{\parbox{\textwidth}{\centering $^1$Computer Graphics Lab, TU Braunschweig, Germany
    \hspace{7pt}{\texttt\small\emph{ \{lastName\}@cg.cs.tu-bs.de}}\\
    \small $^2$ Max Planck Institute for Informatics, Saarland Informatics Campus, Germany
    \hspace{7pt}{\texttt\small\emph{ \{lastName\}@mpi-inf.mpg.de}}
    }
  }
}
\begin{document}

 \teaser{
 \vspace{-1.0cm}
  \includegraphics[width=.9\linewidth, keepaspectratio]{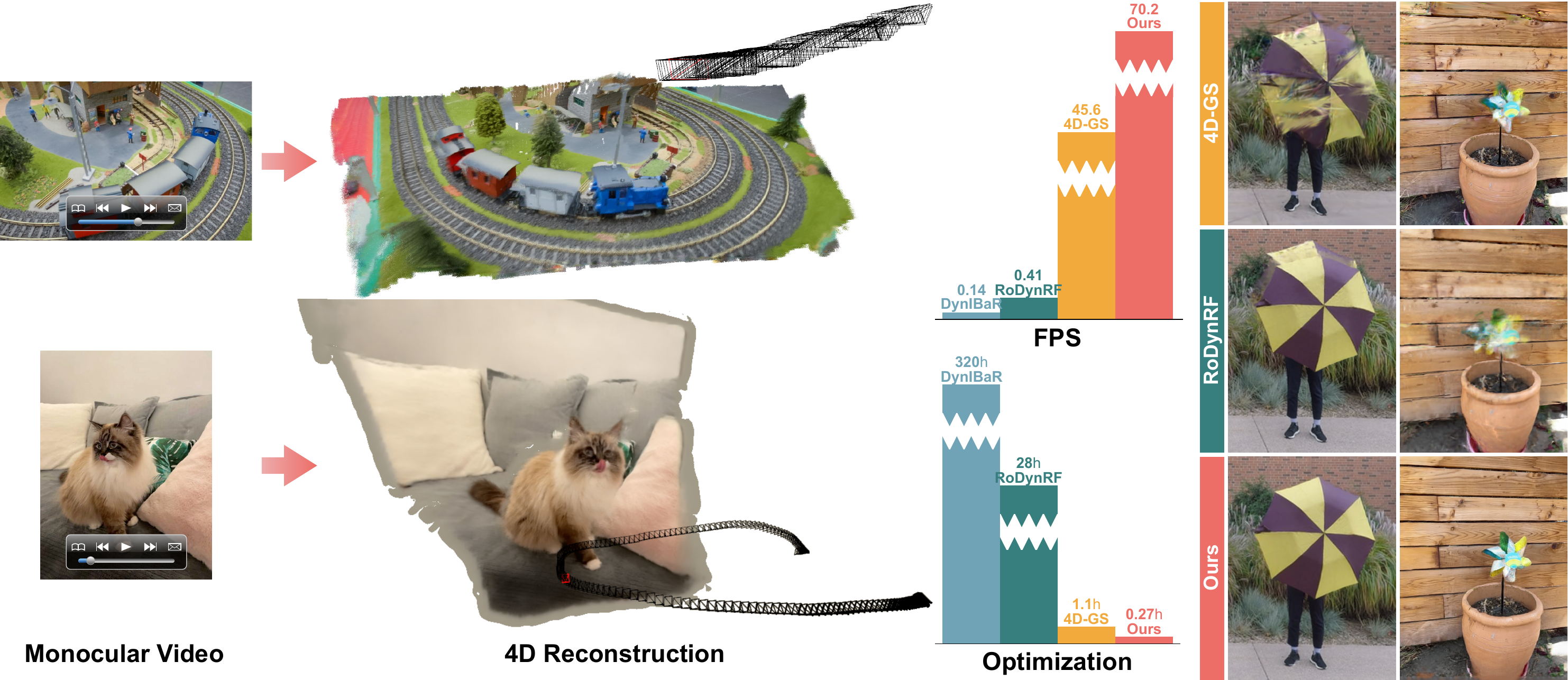} 
  \centering
   \caption{We introduce \textbf{Dynamic Neural Point Clouds} for dynamic view synthesis from monocular videos. Our method enables fast optimization, real-time frame rates, and competitive image quality.}
 \label{fig:teaser}
}

\maketitle

\begin{abstract}
Dynamic reconstruction and spatiotemporal novel-view synthesis of non-rigidly deforming scenes recently gained increased attention.
While existing work achieves impressive quality and performance on multi-view or teleporting camera setups, most methods fail to efficiently and faithfully recover motion and appearance from casual monocular captures.
This paper contributes to the field by introducing a new method for dynamic novel view synthesis from monocular video, such as casual smartphone captures.
Our approach represents the scene as a \textit{dynamic neural point cloud}, an implicit time-conditioned point distribution that encodes local geometry and appearance in separate hash-encoded neural feature grids for static and dynamic regions.
By sampling a discrete point cloud from our model, we can efficiently render high-quality novel views using a fast differentiable rasterizer and neural rendering network.
Similar to recent work, we leverage advances in neural scene analysis by incorporating data-driven priors like monocular depth estimation and object segmentation to resolve motion and depth ambiguities originating from the monocular captures.
In addition to guiding the optimization process, we show that these priors can be exploited to explicitly initialize our scene representation to drastically improve optimization speed and final image quality.
As evidenced by our experimental evaluation, our dynamic point cloud model not only enables fast optimization and real-time frame rates for interactive applications, but also achieves competitive image quality on monocular benchmark sequences.

Our code and data are available online {\small\url{https://moritzkappel.github.io/projects/dnpc/}}.


\end{abstract}  

\section{Introduction}
\label{sec:introduction}
Synthesizing novel views from a sparse set of input images is a fundamental challenge in computer vision. 
Using recent advances in 3D neural scene reconstruction, it becomes possible to create photorealistic renderings from arbitrary viewpoints, effectively adding interactive six degrees-of-freedom camera support.
In the context of general, non-rigidly deforming dynamic scenes, the capability of interactively rendering novel views for every point in time enables numerous applications, including video stabilization~\cite{liu2021hybrid}, immersive playback for social media~\cite{attal2023hyperreel}, or virtual reality~\cite{VRNeRF}.
However, performing faithful reconstruction from just a single monocular video (e.g., a casual smartphone recording) remains a computationally expensive and challenging problem.
While the introduction of Neural Radiance Fields~\cite{mildenhall2021nerf} and 3D Gaussian Splatting~\cite{kerbl20233d} brought about significant advances in static scene reconstruction, dynamic extensions still lag behind in terms of quality and robustness due to motion and depth ambiguities in the temporal domain.
Those ambiguities are usually resolved using expensive large-scale multi view setups~\cite{li2022neural} or monocularized data (i.e., teleporting camera setups)~\cite{kappel2024fast}, which maximize the ratio between camera and scene motion to improve temporal consistency.
However, recent investigations~\cite{gao2022Neurips} reveal that most sparsely regularized methods fail to faithfully recover consistent motion and geometry from casual smartphone captures due to the lack of effective multi-view signal.
Multiple recent radiance field approaches address this problem by incorporating strong priors like depth, optical flow, and semantic segmentation for the joint reconstruction of geometry, 4D scene flow fields~\cite{li2021neural, li2023CVPR}, and even camera parameters~\cite{liu2023CVPR} from monocular video.
While current monocular methods achieve high-quality spatio-temporally consistent view synthesis, they are often limited in the expressiveness of novel camera views, and by their high computational demand.
To address the above-mentioned challenges, this paper introduces \textit{Dynamic Neural Point Clouds} (D-NPC), a new system for monocular dynamic view synthesis based on implicit point distributions and fast hash-encoded feature grids.
Given a monocular input video and the corresponding structure-from-motion calibration, our method optimizes a 4D representation of the depicted scene, which enables spatio-temporal novel-view synthesis, e.g., the infamous bullet-time effect.
Following the concept of implicit neural point clouds~\cite{hahlbohm2024inpc}, we track expected scene density in a sparse voxel grid, which is used to extract explicit pose-dependent point clouds.
Combined with a fast multi-resolution hash-encoded feature grid~\cite{muller2022instant}, this scene representation is able to efficiently render high-fidelity novel views using a differentiable rasterizer and image-to-image translation network.
Based on the concepts of space-time hash encoding~\cite{wang2024masked}, we lift the representation to the temporal domain by replacing the single feature grid with a 3D and 4D hash encoding for the static background and dynamic foreground regions, respectively.
We further extend the point distribution field to robustly track the per-voxel ratio of static and dynamic scene content, effectively enabling 3D dynamics segmentation to significantly reduce the required compute during optimization and inference.
To guide the severely under-constrained monocular reconstruction, we leverage per-frame data-driven priors, namely monocular depth and foreground segmentation masks, which we extract and preprocess with minimal overhead.
We demonstrate that our temporal extension can take advantage of the rasterization-based forward rendering of implicit neural point clouds, which enables efficient foreground-background decomposition and image-space regularization.
Furthermore, the scene-level sampling approach also provides a seamless way of integrating data-driven priors for model initialization, resulting in faster and better convergence during optimization.
As shown in \cref{fig:teaser}, \textit{D-NPC} achieves competitive image quality while significantly reducing optimization time, and allowing for interactive rendering in a graphical user interface.

Overall, we summarize our main technical contributions as follows: 
\begin{itemize}
    \item We propose a new, efficient dynamic scene representation for non-rigid scene reconstruction and novel-view synthesis from a single monocular input video. \textit{D-NPC} is the first adaptation of implicit neural point clouds for dynamic view synthesis.
    \item We demonstrate that our model can elegantly leverage additional priors like monocular depth and object segmentation masks to explicitly initialize the sampling distribution field, which drastically improves optimization and inference speed.
\end{itemize}
Our method enables fast optimization in as few as 0.27 GPU hours and interactive frame rates 
above 60 FPS during inference, while achieving competitive image quality on common benchmark datasets. 
\section{Related Work}
\label{sec:related_work}
Rendering novel viewpoints of a recorded scene is a longstanding and well-studied problem. 
Early work on image-based rendering uses morphing to generate in-between views from image collections without explicit 3D reconstruction~\cite{chen1993view, seitz1996view}.
Other approaches apply implicit or explicit 3D scene representations like light fields~\cite{gortler2023lumigraph}, per-view meshes~\cite{R5-TOG18}, geometric scaffolds~\cite{R5-CVPR21}, or layered depth images~\cite{shade1998layered} to generate high-quality novel views.
%
%
Recently, Neural Radiance Fields (NeRFs)~\cite{mildenhall2021nerf} and 3D Gaussian Splatting (3D-GS)~\cite{kerbl20233d} inspired a multitude of follow-up research, including extensions for dynamic scenes.
We next briefly summarize neural representations for static scenes, followed by dynamic view synthesis.
\subsection{Static View Synthesis}
NeRFs~\cite{mildenhall2021nerf} implicitly represent the scene geometry and appearance using a Multi-Layer Perceptron (MLP), yielding an unprecedented image quality for novel views at the cost of slow optimization and rendering due to the large number of network queries required.
While some follow-ups propose qualitative improvements for, \eg anti-aliasing and unbounded scenes~\cite{barron2021mip, barron2022mipnerf360, barron2023zip}, another branch of methods focuses on acceleration using explicit scene representations like sparse~\cite{fridovich2022plenoxels} or decomposed~\cite{chen2022tensorf} voxel grids.
Along the same lines, Müller~\etal~\cite{muller2022instant} propose an efficient grid-based scene representation implemented as a multi-resolution hash-grid, enabling high-quality reconstruction in only a few minutes.
All the above volume rendering approaches apply \textit{backward rendering}, where the scene is frequently sampled along the camera rays.
In contrast, \textit{forward rendering} approaches, explicitly model scene geometry using geometric primitives, allowing for fast rasterization-based rendering.
The popular 3D-GS by Kerbl~\etal~\cite{kerbl20233d} represents the scene as a set of 3D Gaussians for very fast rendering at high image quality, inspiring many extensions~\cite{lee2023compact, yu2023mip}.
Analogously, point-based rendering approaches~\cite{aliev2020neural, kopanas2021point, ruckert2022adop} use a fixed initial point cloud combined with a neural image translation network for hole filling.
The recent INPC approach by Hahlbohm~\etal~\cite{hahlbohm2024inpc} combines the benefits of forward and backward rendering, modelling the scene as a point probability and hash-encoded appearance field for high-quality rendering at interactive frame rates.
Our \textit{D-NPC} method leverages these benefits for fast dynamic view synthesis from monocular video.
A more detailed discussion of neural rendering techniques can
be found in the report by Tewari~\etal~\cite{Tewari2022NeuRendSTAR}.
\subsection{Dynamic View Synthesis}
Most NeRF-based extensions use a time-conditioned MLP~\cite{li2022neural, li2021neural, park2023temporal} or additional deformation field~\cite{fang2022fast, park2021nerfies, park2021hypernerf, pumarola2021d} to represent scene dynamics.
Following acceleration models for static reconstruction, similar models based on factorized~\cite{attal2023hyperreel, cao2023hexplane, fridovich2023k, wang2023mixed} and hash-encoded~\cite{kappel2024fast, song2023nerfplayer, wang2024masked} feature grids were introduced for dynamic view synthesis.
D\textsuperscript{2}NeRF~\cite{wu2022d2nerf} combined two models for foreground-background separation from a single video.
Likewise, Wang~\etal~\cite{wang2024masked} decompose the scene in a fast 3D and 4D hash-grid representation.
We adapt concepts from both approaches, integrating foreground/background handling in both our sampling and appearance modelling, enabling our method to individually render dynamic and static scene content while improving performance.
Similar to radiance fields, recent approaches opt to extend 3D-GS for non-rigid scenes, tracking the 3D motion of Gaussians over time~\cite{luiten2023dynamic, wu20234d} while retaining fast rendering.
However, as recently revealed by Gao~\etal~\cite{gao2022Neurips}, these methods require multi-view or monocularized~\cite{kappel2024fast} capturing setups to exploit observations from multiple views.
On the other hand, multiple methods focus on dynamic view synthesis from a single monocular input video, integrating additional priors like dense scene flow estimation to resolve motion and depth ambiguities~\cite{li2021neural, li2023CVPR, liu2023CVPR}.
RoDynRF~\cite{liu2023CVPR} further takes the challenging scenario to the extreme, additionally optimizing camera poses from the unposed input video.
However, these MLP-based monocular approaches rely on backward rendering, thus being limited in optimization speed and final rendering frame rate.
To circumvent slow optimization, the PGDVS approach by Zhao~\etal~\cite{Zhao2024PGDVS} proposes a generalized approach for monocular video that bypasses per scene optimization.
Still, the slow inference speed and varying generalization quality largely rule out application for real-time downstream tasks.
To jointly enable fast high quality per-scene optimization and subsequent real-time novel-view synthesis, this paper presents a point-based forward rendering approach for monocular view synthesis, which leverages additional priors to initialize and guide the reconstruction.
More details on the state of the art in dynamic reconstruction can be found in the survey by Yunus~\etal~\cite{yunus2024recent}.
\subsection{Concurrent Work}
Concurrent to our work, other approaches propose fast forward rendering representations for dynamic reconstruction and novel-view synthesis from monocular video.
Please note that the covered concurrent methods are not yet published in any proceedings (at the time of submission), or constitute non-peer reviewed work on arXiv.
Similar to our approach, Lee~\etal~\cite{lee2023casual-fvs} separately model the foreground and background, but instead use a plane based proxy geometry for static regions, and points for the dynamic foreground, enabling fast forward rendering at a decent image quality.
Following up on the adaption of 3DGS for dynamic multi-view recordings, Stearns~\etal~\cite{stearns2024dynamic} propose a divide-and-conquer approach for motion and appearance reconstruction from single monocular video using isotropic Gaussians.
Like most Gaussian-based methods, this approach inherits impressive online framerates and decent image quality, but adds overhead in terms of optimization time.
The Shape of Motion approach by Wang~\etal~\cite{som2024} additionally uses 2D motion track inputs to recover 3D scene motion represented via decomposed motion bases, which is applied to a set of canonical 3D Gaussians for novel view synthesis.
Lei~\etal~\cite{mosca2024} also propose to disentangle the deformations of the scene from its geometry. The former is represented as a sparse graph of motion trajectories that they lift from 2D priors and optimize exploiting semantic features and the assumption of the deformations being as-rigid-as-possible. They use Gaussian Splatting to render the scene resulting from the global fusion of the 3DGs per timestep and deformed accordingly to the motion graph.
While our method also involves rendering explicit geometry for improved performance, using an implicit point cloud representation instead of explicit Planes, Points, or Gaussians entails unique characteristics regarding performance, quality and robustness.

\section{Method}
\label{sec:method}
\begin{figure*}[ht]
    \includegraphics[width=1.0\textwidth, keepaspectratio]{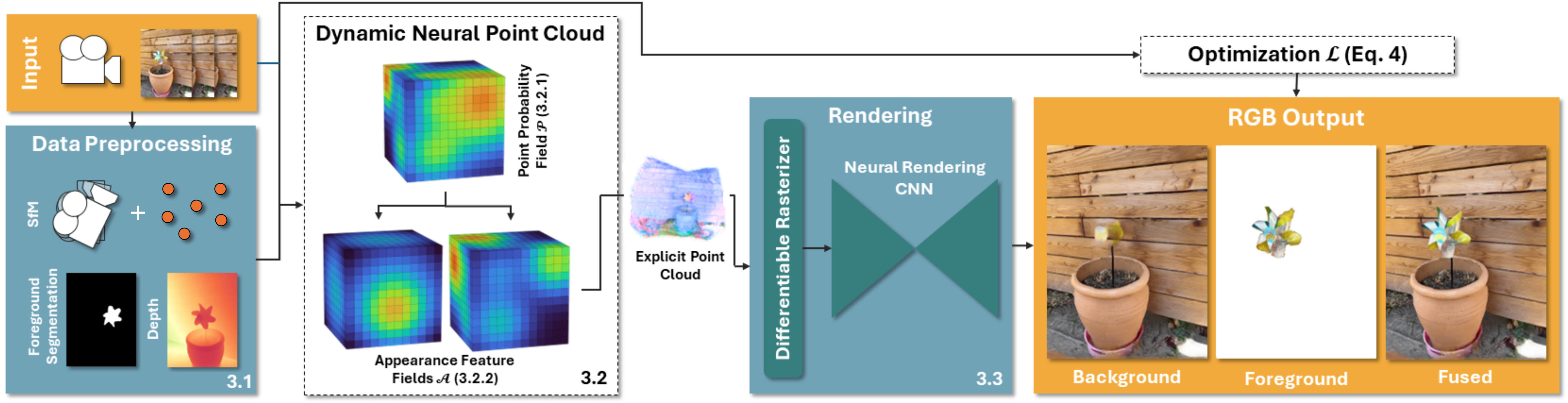} 
    \caption{%
        \textbf{Method overview}. Given a monocular RGB video and priors extracted from out-of-the-box estimators, we initialize and optimize a dynamic implicit neural point cloud, consisting of a spatiotemporal point position distribution and two feature grids for static and dynamic scene content. By sampling an explicit point cloud for a discrete timestamp, our model can synthesize novel views, including foreground/background separation, using a differentiable rasterizer and neural renderer.%
    }\label{fig:framework} 
\end{figure*} 
We introduce \textit{Dynamic Neural Point Clouds (D-NPC)}, a method for dynamic novel-view synthesis from monocular video.
Given a single monocular video consisting of $N$ images $I = (i_1, ..., i_N)$ and their normalized timestamps $T = (t_1, ..., t_N) \in [0, 1]^{N}$, we reconstruct a dynamic implicit point cloud representation~\cite{hahlbohm2024inpc} of the depicted dynamic scene.
Based on its static counterpart, the core of our \textit{D-NPC} model consists of a spatiotemporal point probability ($\mathcal{P}$) and appearance feature descriptor ($\mathcal{A}$) field, which can efficiently be rendered using a differentiable rasterizer and neural rendering network.
A full overview of our method is given in \cref{fig:framework}.
In the following, we describe the individual components and optimization process in detail.
\subsection{Data Preprocessing}
\label{sec:data}
The reconstruction of non-rigidly deforming scenes from a single moving camera is a highly ill-posed problem.
To regularize the reconstruction from monocular video, previous approaches~\cite{gao2022Neurips, li2021neural, liu2023CVPR} apply estimates from several image processing applications, replacing the multi-view consistency signal with learned, data-driven scene analysis. 
Following this idea, we first extract additional priors, which we use to initialize and guide our model in the later stages.
Despite the recent emergence of consistent yet expensive scene calibration~\cite{smith24flowmap}, video depth estimation~\cite{Luo-VideoDepth-2020} and motion tracking~\cite{wang2023omnimotion} estimators, one of our method's main objectives is speed and simplicity.
Thus, we focus on fast out-of-the-box solutions to maintain a swift total reconstruction time (including prior computation).
First, we start by applying sparse structure-from-motion (SfM) using COLMAP~\cite{schoenberger2016sfm, schoenberger2016mvs} or VGGSfM~\cite{wang2023vggsfm} to estimate camera intrinsics and per-frame poses (extrinsics) $E = (e_1, ..., e_N)$.
As a side product, we extract the sparse, globally-aligned SfM point cloud.
To reduce the depth ambiguity of dynamic scene content, we employ the fast monocular depth estimator DepthAnything~\cite{yang2024depth}, which provides relative disparity maps for every video frame.
Instead of using a relative depth loss~\cite{li2021neural}, we align the predicted disparity with the global SfM point cloud by estimating a scale and shift via RANSAC-based least squares and temporal smoothing.
This way, we obtain globally-aligned depth which can \textit{explicitly} be integrated into our model.
Alternatively, our method also accepts (sparse) metric LiDAR depth and camera poses obtained, e.g., via IMUs, which completely lifts the dependency on SfM calibration and point cloud priors.
Lastly, we obtain binary segmentation masks $\omega$ of moving foreground object(s) to mask out dynamic scene content in our reconstruction and COLMAP SfM feature extraction.
Similar to monocular depth, automatic motion segmentation is highly ill-posed, even in the presence of semantic segmentation masks.
Thus, we use the semi-automatic video object segmentation system Cutie~\cite{cheng2023putting}, which is guided by minimal user input.
In line with our demands on simplicity, Cutie generates masks of sufficient quality in less than a minute, requiring as few as two mouse clicks, which in the future could be integrated into the data recording setup on mobile devices.
\subsection{Dynamic Neural Point Cloud Representation}
Our method represents the 4D scene as a Dynamic Neural Point Cloud, the first temporal extension of implicit neural point clouds~\cite{hahlbohm2024inpc}.
In their paper, Hahlbohm~\etal\ propose a methodology that diverges from the conventional approach of storing both the spatial and photometric properties of the scene within a unified data structure. Instead, they advocate for a division of these properties and an implicit optimization of both. The geometric structure is depicted as an octree-based point probability field ($\mathcal{P}$), which undergoes progressive subdivision to ensure a uniform probability distribution across all leaf nodes. Conversely, the appearance features are embedded in an implicit coordinate-based multi-resolution hash grid ($\mathcal{A}$). During rendering, $\mathcal{P}$ is employed as an estimator for point positions, utilizing either random positions in each leaf or fixed sampling patterns. Meanwhile, per-point appearance features are queried from $\mathcal{A}$. 
The authors then employ fast bilinear splatting to render the resulting explicit point cloud, where gradients are backpropagated through their pipeline to the implicit representation~\cite{hahlbohm2024inpc}.

Thus, the primary idea of implicit point clouds is to replace the persistent point cloud $(p, f)$, consisting of a set of 3D positions $p$ and corresponding color or feature descriptor vectors $f$, with a sampling distribution $\mathcal{P}$ and a neural feature grid $\mathcal{A}$.
By repeatedly sampling and rendering an explicit point cloud $((p, \mathcal{A}(p)), p \sim \mathcal{P})$, it is possible to iteratively update and refine the spatial distribution, while maintaining all benefits of point-based rendering.
However, applying this concept to dynamic scenes is not straightforward, as the direct integration of a fourth dimension is memory-intensive and does not naturally exploit any temporal coherency.
\subsubsection{Temporal Point Probability Field}
The probability field $\mathcal{P}$ models the likelihood for spatial point sampling by tracking the estimated local scene opacity $\sigma \in [0, 1]$ over the course of optimization.
To facilitate accurate and efficient sampling over a dynamic sequence, we divide our probability field $\mathcal{P} = (\mathcal{P}_s, \mathcal{P}_d)$ into two separate instances for static and dynamic regions, respectively.
While the static 3D field $\mathcal{P}_s$ continues to track local opacity $\sigma_s$ assumed to be consistent over all points in time, the 4D dynamic field $\mathcal{P}_d$ stores additional values $\sigma_{d}(t)$ as a function of time.
To enable a clean separation of foreground and background, $\mathcal{P}_d$ not only models the expected local opacity, but additionally keeps track of the estimated local proportion $\beta(t)$ between static and dynamic substance, as explained later on.
In conjunction, our probability fields can sample explicit, time-aware point positions, while pre-marking individual points as either dynamic or static to reduce compute complexity.
Given the natural imprecisions and relatively low resolution of monocular reconstruction, we find that the octree-based data structure suggested by Hahlbohm~\etal\ does not yield significant benefits due to the absence of fine geometric details.
Instead, we use a fixed-resolution sparse voxel grid representation for the static and dynamic fields.
\paragraph{Voxel Grid Initialization.}
Despite using a sparse data structure, starting from a fully occupied grid (i.e., all cells are sampled equally for all timestamps) is extremely inefficient regarding both memory consumption and model convergence.
Thus, we make use of the data-driven priors described in \cref{sec:data} to find a sparse initial sampling distribution before optimization.
First, we initialize the voxel grid bounds as the minimal cuboid enclosing all SfM points after performing outlier filtering based on the 95-percentile, containing a fixed total of $128^{3}$ and $N \cdot 128^{3}$ voxels for $\mathcal{P}_s$ and $\mathcal{P}_d$ respectively.
Then, for every timestamp $t_{i} \in T$, we use the corresponding training camera matrix $e_i$ to project a set of random points in each voxel to the image plane, and gather the minimal relative distance to the estimated monocular depth $d_i$, and the maximum over the binary foreground segmentation mask $\omega_i$.
For the dynamic probability field $\mathcal{P}_d$, we set the initial opacity-based sampling probability $\sigma_d(t_{i})$ of each cell based on the relative deviation ($d\in [0,1]$) from the monocular depth: $\sigma_d(t_{i})=(1-d(t_{i}))^{x}$ with $x$ being a free hyperparameter, to initially sample more points close to the expected surface. For the static probability field ($\mathcal{P}_s$), we instead use the maximum over all timestamps ($\sigma_{s} = \max\{\sigma_{d}(t_i)\:|\: t_i \in T\}$).

Additionally, we carve the temporal slices of $\mathcal{P}_d$ using the dynamic segmentation.
By thresholding the sampling probabilities, we eliminate up to 95\% of voxels, resulting in an efficiently represented distribution constituting a proper local minimum for optimization.
\paragraph{Explicit Point Cloud Sampling.}
\label{sec:sampling}
For an arbitrary camera view $e$ and timestamp $t$, we sample a set $P \in \mathbb{R}^{k \times 3}$ of $k$ explicit point positions ($P \sim \mathcal{P} \: | \: (e, t)$) for rendering a novel view.
To this end, we first take the union of all voxels in the static field and the dynamic field at time $t$, taking the maximum expected opacity of voxels existing in both models.
Then, we eliminate voxels based on the camera frustum by projecting each voxel center to the image plane using $e$.
To obtain the final point positions $p$, we interpret the per-voxel opacity as a discrete PDF, using discrete multinomial sampling (with replacement) to propose a list of voxels, in each of which we place a single uniformly sampled 3D position.
This way, dense point clouds are sampled in clusters of high expected scene density, while empty or homogeneous static areas are sampled sparsely, enabling the renderer to infer depth ordering and occlusions via alpha-blending.
For the resulting set of point positions, we further trace a binary mask $m_p$ indicating if a point was sampled from static or dynamic scene regions, enabling efficient model evaluation  by omitting excessive queries of the dynamic appearance model.  
\paragraph{Voxel Grid Update.}
\label{sec:update}
We update the sampling distribution field by tracking the blending weights $(b_i \in \mathcal{B})$ assigned to every sampled explicit point during rasterization.
We first apply a constant decay factor $(\gamma)$ to the values of every unique voxel proposed by the multinomial sampling, and then assign it the maximum over the current value and the blending weights of all points sampled from the specific cell:
\begin{equation}
\sigma_{s} = \max\{\sigma_{s}\gamma,\ max\{\mathcal{B}\}\}, 
\end{equation}
where $\mathcal{B}$ is the set of blending weights, i.e., the visibility of all points sampled from this cell, which is calculated during alpha-blending at the rasterization step.
For samples from dynamic voxels of $\mathcal{P}_d$, we apply the same procedure to the tracked static-to-dynamic ratios $\delta$ using per-point densities described in \cref{sec:appearance_field}.
Afterwards, we prune voxels from the sampling distribution by threshold:
For $\mathcal{P}_d$, we remove cells with a low sampling probability, as it either describes empty space or is well handled by the static model.
We further remove static voxels from $\mathcal{P}_s$ whose expected opacity $\sigma$ is low both in the static 3D field and all timestamps in the dynamic field.
This way, we progressively reduce memory consumption and refine importance sampling.
\subsubsection{Dynamic Appearance Field}
\label{sec:appearance_field}
In a conventional point cloud, every point is assigned a fixed feature vector describing its appearance.
As implicit neural point clouds~\cite{hahlbohm2024inpc} repeatedly extract new sets of point positions, they use a fast multi-resolution hash-encoded feature grid~\cite{muller2022instant} $\mathcal{A}$ to model the appearance over the entire spatial sampling distribution.
To represent time-conditioned appearance for dynamic scenes, we again use a static 3D and dynamic 4D hash-encoded neural feature grid $\mathcal{A} = (\mathcal{A}_{s}, \mathcal{A}_{d})$.
As previously described by Wang~\etal~\cite{wang2024masked}, using an ensemble of 3D and 4D hash-grids instead of a single 4D representation effectively improves the capabilities of the dynamic model by reducing hash collisions.
To model the static and dynamic scene appearance at a spatial input position $p$ at time $t$, the appearance fields yield a density $\delta \in \mathbb{R}_{0}^{+}$ and an $n_f$-dimensional feature vector $f\in [0, 1]^{n_{f}}$:
\begin{equation}
    \delta_{s}, f_{s} = \mathcal{A}_{s}(p), \quad \delta_{d}, f_{d} = \mathcal{A}_{d}(p, t).
\end{equation}
Note that we only have to query $\mathcal{A}_d$ for points indicated as dynamic by the binary segmentation mask $m$, which significantly reduces compute and memory.
Following the formulation of D\textsuperscript{2}NeRF~\cite{wu2022d2nerf}, we then calculate the combined per-point density $\delta_{c}$ as the sum of static and dynamic densities, and obtain the final feature descriptor $f_{c}$ via density-weighted interpolation:
\begin{equation}
    \label{eq:fusing}
    \delta_{c} = \begin{cases}\delta_{s} + \delta_{d} & \text{if} \quad m(p) \\ \delta_{s} & \text{else} \end{cases}, 
    \qquad f_{c} = \begin{cases}\frac{\delta_{s}f_{s} + \delta_{d}f_{d}}{\delta_{c}} & \text{if} \quad m(p) \\ f_{s} & \text{else} \end{cases}
\end{equation}
For rendering and probability field updates, we then transform density values $\delta$ into opacity $\sigma$, and calculate the dynamic ratio $\beta$ via:
\begin{equation}
    \sigma_{(s/d/c)} = 1 - e^{-\delta_{(s/d/c)}}, \quad \beta = \frac{\delta_d}{\delta_c}.
\end{equation}

\subsection{Neural Point Cloud Rendering}
\label{sec:render}
Given the time-dependent neural point cloud for an input view consisting of point positions $p$, point opacity $\sigma$, and appearance features $f$, we implement a fast differentiable point rasterizer to generate alpha ($\alpha$), depth ($d$), and 2D image feature maps $f_{img}$.
Naturally, the rasterized images still contain holes, a common characteristic of classical point rendering approaches.
Thus, we use a neural renderer, implemented as a light-weight convolutional neural network with a U-Net~\cite{ronneberger2015u} architecture. This network translates the sparse rasterized feature image ($f_{img}$) to a dense output image ($c$), thus inferring human-interpretable RBG color from the higher dimension point cloud features while filling the remaining holes based on neighborhood information.
In practice, we apply the rasterization and rendering components to the static, dynamic, and combined point cloud features separately, resulting in individual depth $d_{s, d, c}$ and color $c_{s, d, c}$ images that can be used for supervision and regularization.
As the resolution of dynamic view synthesis benchmarks is usually low, we do not use any splatting or multi-resolution inputs in the rasterizer or neural renderer, which further increases performance.
\subsection{Optimization}
\label{sec:optim}
During training, we iteratively sample an explicit point cloud (\cref{sec:sampling}) for a random training view and its associated timestamp, extract per-point appearance features (\cref{sec:appearance_field}), render the resulting images (\cref{sec:render}), update the sampling distribution (\cref{sec:update}), and optimize model parameters using the following objective function:
\begin{equation}
    \mathcal{L} = \lambda_{\text{photo}}\mathcal{L}_{\text{photo}} + \lambda_{\text{depth}}\mathcal{L}_{\text{depth}} + \lambda_{\text{seg}}\mathcal{L}_{\text{seg}} +
    \lambda_{\text{dist}}\mathcal{L}_{\text{dist}},
\end{equation}
with hyperparameters $\lambda_{\text{photo}}$, $\lambda_{\text{depth}}$, $\lambda_{\text{seg}}$ and $\lambda_{\text{dist}}$. We use the same set of hyperparameters across all our experiments. 
The loss $\mathcal{L}_{\text{photo}}$ enforces photometric consistency between the rendered ($c$) and GT RGB image.
Here, we use a weighted combination of pixel-level Cauchy loss~\cite{black1996robust}, structural dissimilarity~\cite{wang2004image}, and perceptual LPIPS~\cite{zhang2018unreasonable} to maximize image quality.
Our depth regularization loss $\mathcal{L}_{\text{depth}}$ penalizes the difference between the rasterized dynamic depth $d_d$ and the aligned monocular depth prior $d_{\text{gt}}$ in the binary foreground mask $\omega$:
\begin{equation}
    \mathcal{L}_{\text{depth}} = \frac{1}{\lvert\omega \rvert} \sum_{i=0}^{\lvert\omega\rvert}\omega_{i}\lvert d_{d, i} - d_{\text{gt}, i}\rvert
\end{equation}
The segmentation regularizer $\mathcal{L}_{seg}$ encourages a clean separation between the static and dynamic models via skewed binary entropy~\cite{wu2022d2nerf} on the dynamic ratios $\beta$:
\begin{equation}
    \mathcal{L}_{\text{seg}} = -\beta^{k}\log_{2}(\beta^{k}) - (1-\beta^{k})\log_{2}(1-\beta^{k}),
\end{equation}
where $k$ is the skewing bias.
As our last regularizer $\mathcal{L}_{dist}$, we adapt the distortion loss of Mip-NeRF360~\cite{barron2022mipnerf360, sun2022improved} to encourage thin surfaces, as it is not feasible to reliably reconstruct the time-dependent volume of a deforming object from a single monocular observation.
For a detailed description of our training and loss schedule, please see our supplemental material.
\begin{table}[thb]
    \centering
    \scriptsize{ 
    \setlength\tabcolsep{3pt}
    \begin{tabular}{@{}l c c c}
        \toprule
        \textbf{Method} & \textbf{GPU Hours [h]} $\downarrow$  & \textbf{Frame Rate [FPS]} $\uparrow$ & \textbf{Frame Rate [FPS]} $\uparrow$\\
        & (NVIDIA dataset) & ($480{\times}270$) & ($1920{\times}1080$)\\
        \midrule
        RoDynRF~\cite{liu2023CVPR}$^\dagger$ & 28 & ${<}1$ & --\\
        HyperNeRF~\cite{park2021hypernerf}$^\dagger$ & 64 & ${<}1$ & --\\
        DynamicNeRF~\cite{gao2021dynamic}$^\dagger$ & 74 & ${<}0.1$ & --\\
        NSFF~\cite{li2021neural}$^\dagger$ & 223 & ${<}1$ & --\\
        DynIBaR~\cite{li2023CVPR}$^\dagger$ & 320 & ${<}1$ & --\\
        \midrule
        Marbles~\cite{stearns2024dynamic} $^\dagger$ & 3.5 & 200+ & -- \\
        CasualFVS~\cite{lee2023casual-fvs} $^\dagger$ & 0.25 & 48 & -- \\
        4D-GS~\cite{wu20234d} & 1.1 & 45 & 30 \\
        \midrule
        \textbf{Ours} & 0.27 & 78 & 22 \\
        \bottomrule
        &&&\\[-.1cm] 
        \multicolumn{2}{l}{\textsuperscript{$\dagger$} Values gathered from literature} & \multicolumn{2}{l}{-- Values not reported} \\
    \end{tabular}
    }
 \caption{\textbf{Performance Comparison}. We report training times on the NVIDIA dataset, and inference frame rates for training- and FHD resolution.}%
 \label{tab:timing}
\end{table}
%
\begin{figure*}
    \centering
    \setlength\tabcolsep{0pt}
    \small{
        \begin{tabular}{*{5}{p{0.195\linewidth}<{\centering}}}
            TiNeuVox & RoDynRF & 4D-GS & Ours & Ground Truth
        \end{tabular}
    }
    \includegraphics[width=.99\linewidth, keepaspectratio]{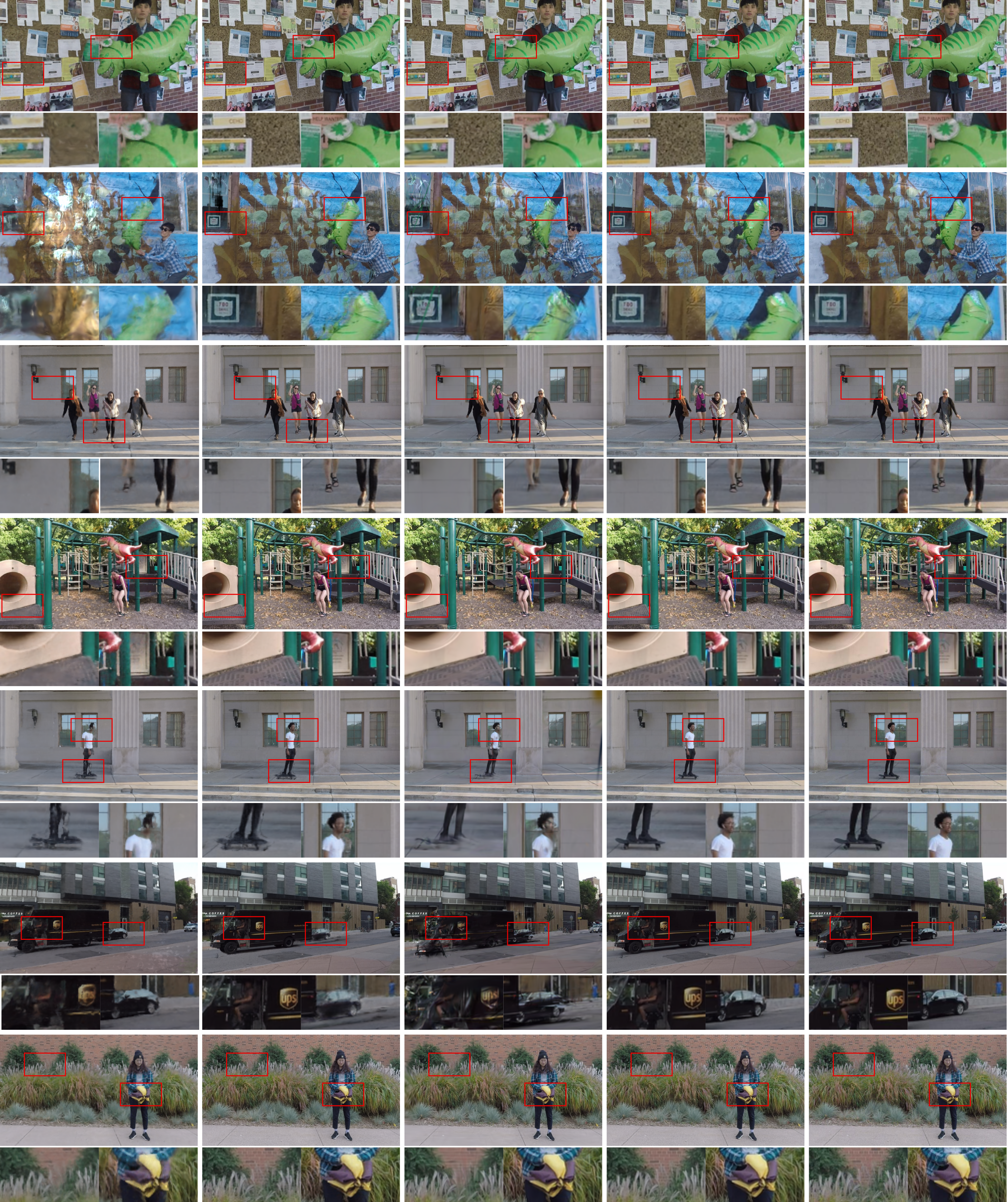}
    \caption{%
        \textbf{Qualitative comparisons on the NVIDIA dataset}. Our method preserves fine details and clean foreground-background transitions.%
    }\label{fig:nvidia}
\end{figure*}
\begin{figure*}
    \centering
    \setlength\tabcolsep{0pt}
    \small{
        \begin{tabular}{*{7}{p{0.16665\linewidth}<{\centering}}}
            HyperNeRF\textsuperscript{$\dagger$} & T-NeRF\textsuperscript{$\dagger$} & RoDynRF & 4D-GS & Ours & Ground Truth
        \end{tabular}
    }
    \includegraphics[width=\linewidth, keepaspectratio]{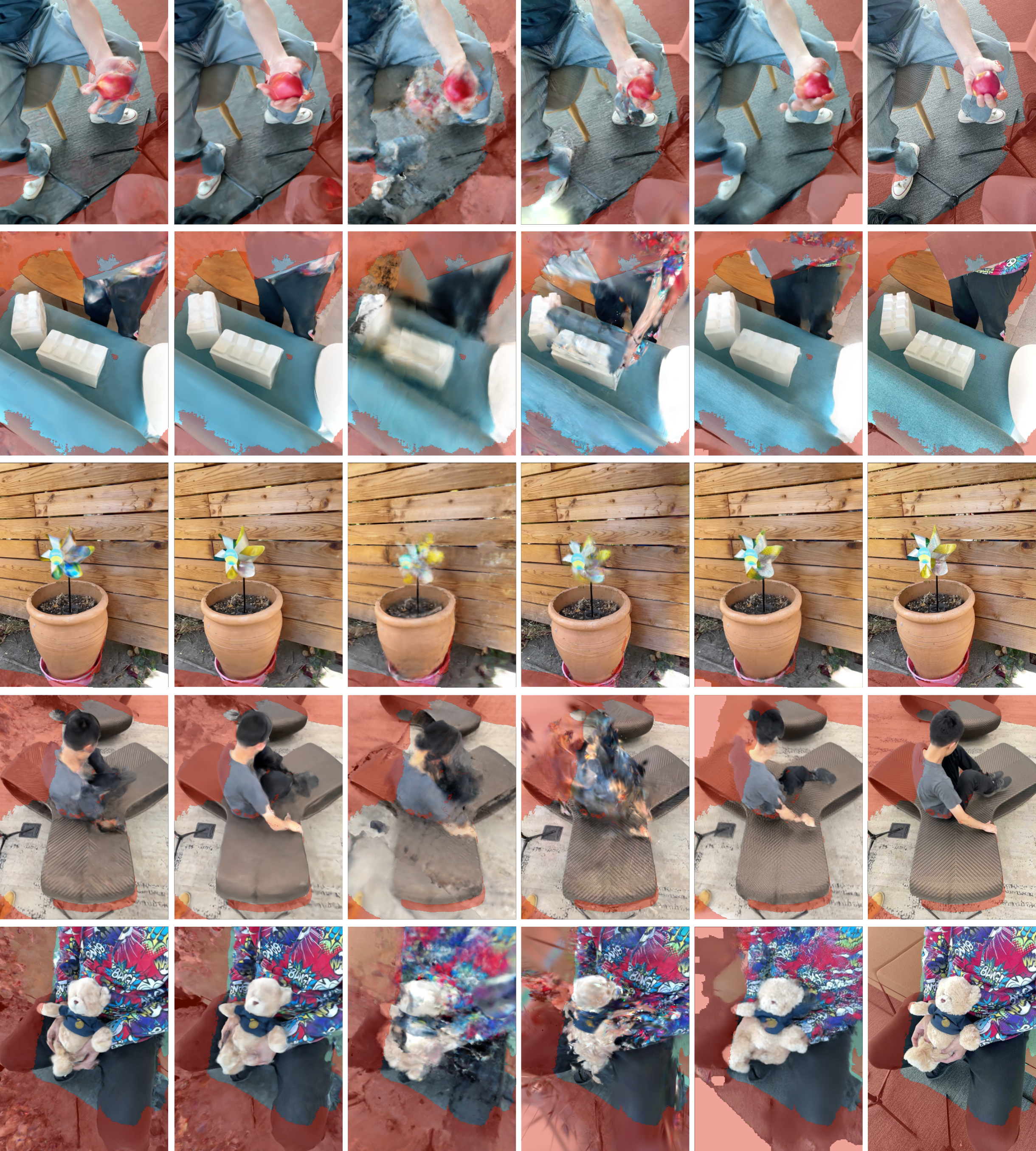}
    \caption{%
        \textbf{Comparison on the iPhone dataset}. We compare on five challenging scenes of the iPhone dataset. Our method better preserves the object shapes and fine details despite complex deformations. Methods indicated by $\dagger$ are the highest quality versions provided by Gao~\etal~\cite{gao2022Neurips}. The areas highlighted in red indicate lack of co-visibility between training and validation.%
    }\label{fig:iphone} 
\end{figure*}

\section{Experimental Evaluation}
\label{sec:experimental_evaluation}
We perform quantitative (\cref{sec:quantitative}) and qualitative (\cref{sec:qualitative}) evaluation and comparisons against state-of-the-art methods on two benchmark datasets:
First, we evaluate our approach on the NVIDIA dynamic scenes dataset~\cite{Nvidia:2020}, featuring forward-facing scenes captured by a multi-view rig, monocularized by using a single frame per time to get a smooth camera trajectory.
We extract training data using the procedure of RoDynRF~\cite{liu2023CVPR},
resulting in $12$ training and test views.
Additionally, we evaluate our approach on the iPhone dataset~\cite{gao2022Neurips} which comprises challenging sequences with complex camera and scene motion trajectories, fine texture details, and (self-)occlusions.
We further show qualitative results including foreground-background separation on high-resolution (1080p) in-the-wild scenes from the DAVIS dataset~\cite{Perazzi2016}, and provide an ablation study assessing the influence of individual pipeline components in \cref{sec:ablation}.
For further experiments, including comparisons against DynIBaR~\cite{li2023CVPR} and results on the HyperNeRF dataset~\cite{park2021hypernerf}, please see our supplemental PDF.
\subsection{Quantitative Results}
\label{sec:quantitative}
We compare the performance of our approach to several related benchmark methods.
In addition to recent NeRF-based approaches, we also compare against the official implementation of 4D-GS~\cite{wu20234d}, a fast state-of-the-art forward rendering approach based on Gaussian splatting.
For both 4D-GS and our method, we measure frame rates using the procedure described by SMERF~\cite{duckworth2023smerf}, rendering the test set $100$ times using a single RTX 4090 GPU.
Please note that we can not report all numbers for concurrent (\textit{unpublished at time of submission}) methods~\cite{lee2023casual-fvs, som2024, stearns2024dynamic, mosca2024} due to missing source code or differences in evaluation procedure (e.g., training resolution or camera calibration on the iPhone dataset).
%
\begin{table*}[th]
    \centering
    \scriptsize{ 
        \setlength\tabcolsep{12pt}
        \begin{tabular}{@{}lccc}
            \toprule
            & \multicolumn{3}{c}{\textbf{NVIDIA Average}} \\
            \textbf{Method} & \textbf{PSNR} $\uparrow$ & \textbf{SSIM} $\uparrow$ & \textbf{LPIPS} $\downarrow$ \\ 
            \midrule
            T-NeRF~\cite{pumarola2021d} &  18.33 & 0.436 & 0.511\\
            D-NeRF~\cite{pumarola2021d} &  21.49 & 0.629 & 0.320\\
            HyperNeRF~\cite{park2021hypernerf} & 17.60 & 0.377 & 0.502\\
            NR-NeRF~\cite{tretschk2021non} & 19.69 & 0.510 & 0.434\\
            NSFF~\cite{li2021neural} & 24.33 & 0.751 & 0.257\\
            DynamicNeRF~\cite{gao2021dynamic} & \cellcolor{1st}26.10& \cellcolor{3rd}0.837& \cellcolor{3rd}0.142\\
            RoDynRF~\cite{liu2023CVPR} & \cellcolor{2nd}25.89& \cellcolor{1st}{0.854}& \cellcolor{2nd}0.110\\
            TiNeuVox~\cite{fang2022fast} & 19.74 & 0.501 & 0.412\\
            4D-GS~\cite{wu20234d} & 22.89 & 0.731 & 0.237\\
            \midrule
            \textbf{Ours} & \cellcolor{3rd}25.64& \cellcolor{2nd}0.845& \cellcolor{1st}0.109\\
            \bottomrule
        \end{tabular}
        \hspace{0.09\textwidth}
        %
        %
        \begin{tabular}{@{}lccc}
            \toprule
            & \multicolumn{3}{c}{\textbf{iPhone Average}} \\
            \textbf{Method} & \textbf{mPSNR} $\uparrow$ & \textbf{mSSIM} $\uparrow$ & \textbf{mLPIPS} $\downarrow$\\ 
            \midrule
            HyperNeRF+B+D+S~\cite{park2021hypernerf} & 16.81 & 0.569 & \cellcolor{2nd}0.332 \\
            Nerfies+B+D+S~\cite{park2021nerfies} & 16.45 & 0.570 & \cellcolor{3rd}0.339 \\
            T-NeRF~\cite{gao2022Neurips} & 15.70 & 0.538 & 0.458 \\
            T-NeRF+B+D~\cite{gao2022Neurips} & \cellcolor{1st}17.06 & 0.573 & 0.390 \\
            T-NeRF+B+D+S~\cite{gao2022Neurips} & \cellcolor{3rd}16.96 & \cellcolor{3rd}0.577 & 0.379 \\
            RoDynRF~\cite{liu2023CVPR} & \cellcolor{2nd}17.00 & 0.535 & 0.517 \\
            \midrule
            Marbles~\cite{stearns2024dynamic} & 16.72 & -- & 0.418 \\
            4D-GS~\cite{wu20234d} & 13.87 & 0.459 & 0.407 \\ 
            \midrule
            \textbf{Ours} & 16.04 &  \cellcolor{2nd}0.589 & 0.348 \\
            \textbf{Ours$_{\text{Lidar}}$} & 16.79 & \cellcolor{1st}0.597 & \cellcolor{1st}0.321\\
            \bottomrule
        \end{tabular}
    }
    \caption{%
        \textbf{Quantitative comparisons on the NVIDIA and iPhone datasets} against recent dynamic view synthesis approaches. Our method achieves the best average scores in terms of perceptual image quality. The three best results are highlighted in \textcolor{1stText}{\textbf{green}} in descending order of saturation.%
    }\label{tab:metrics_combined}
\end{table*}

%
As evident from \cref{tab:timing}, our method is significantly faster than backward rendering approaches, and even trains faster than 4D-GS while achieving similar frame rates during inference.
A breakdown of timings for individual method components is provided in the supplemental material.
We further compare image quality based on PSNR, SSIM~\cite{wang2004image}, and LPIPS~\cite{zhang2018unreasonable} quality metrics on the NVIDIA and iPhone datasets in \cref{tab:metrics_combined}.
To avoid inconsistencies across the reported values, we use the official results and models kindly provided by Liu~\etal~\cite{liu2023CVPR}, Li~\etal~\cite{li2023CVPR}, and Gao~\etal~\cite{gao2022Neurips} to consistently recalculate all metrics, using the VGG backbone for LPIPS calculation.
On the iPhone dataset, we instead use masked metric calculation (mPSNR, mSSIM, mLPIPS) as suggested and implemented by Gao~\etal~\cite{gao2022Neurips} to account for co-visibility.
For our method, we further report values for two versions, trained with our monocular depth alignment (Ours) and the provided Lidar depth (Ours$_{\text{Lidar}}$).
Overall, we find our method is not only the fastest in terms of reconstruction and rendering speed but also achieves competitive image quality, especially in terms of perceptual image quality metrics (LPIPS).
While RoDynRF and T-NeRF achieve better per-pixel PSNR scores, we find that both methods fail to preserve high frequencies, resulting in an inferior perceived image quality, as shown in the next section.
Detailed per-scene results can be found in our supplement. 

\subsection{Qualitative Results}\label{sec:qualitative}
We show qualitative comparisons on the NVIDIA (\cref{fig:nvidia}) and iPhone (\cref{fig:iphone}) datasets.
For the iPhone sequences, we show the best version of HyperNeRF~\cite{park2021hypernerf} and T-NeRF~\cite{pumarola2021d} according to image quality metrics, using all additional regularizers (Lidar depth, random background, and distortion loss) proposed by Gao~\etal~\cite{gao2022Neurips}.
Again, we find that our method achieves a competitive and oftentimes superior perceived image quality, preserving higher frequencies and clean edges.
As our 4D model is temporally conditioned and does not directly estimate inter-frame motion, it can well handle scene movements of varying complexity (e.g., the rotating motion on the Paper-windmill scene in \cref{fig:iphone} or human motion such as in the Skating scene in \cref{{fig:nvidia}}), as long as it is constantly observed from the input camera trajectory (and not occluded).

\begin{figure}[bt]
    \centering
    \includegraphics[width=\columnwidth, keepaspectratio]{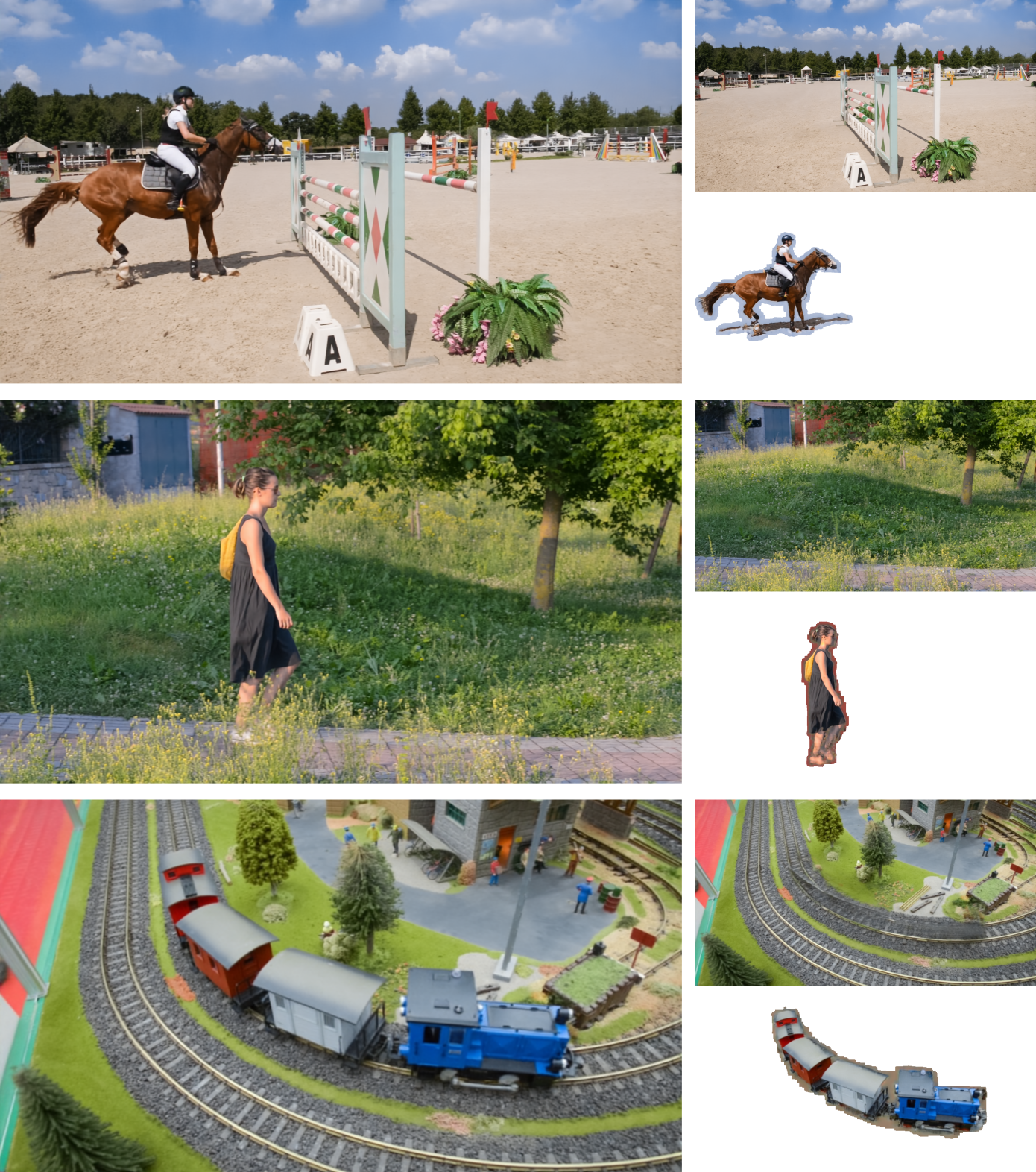}
    \caption{%
        \textbf{Novel views synthesized using complex in-the-wild sequences of the DAVIS dataset.} On the right, we show the corresponding visualizations of our implicit foreground/background separation.%
    }\label{fig:Davis}
\end{figure}
%
\Cref{fig:Davis} further provides novel views synthesized using complex in-the-wild sequences from the DAVIS dataset. 
We additionally include visualizations of our implicit foreground/background separation obtained by only rendering the plain static or dynamic components of the sampled point clouds. 
Given that our method is efficient in both memory and compute, we train our DAVIS models on the full original resolution (FHD) and double the sampled point cloud size to $8M$ points to maintain a dense distribution in the rasterized feature images.
This upgrade in rendering resolution and point queries increases the training time to $0.52$ hours and reduces online framerates to $18$ fps.
Overall, our experiments show that DNPC robustly handles high-resolution in-the-wild inputs provided that decent camera poses can be obtained from current SfM methods.
For more video results, including foreground/background separation, 3D anaglyphs renderings and live GUI capture, please see our supplemental material.
\subsection{Ablation Study}
\label{sec:ablation}
\begin{table}[thb]
    \centering
    \scriptsize{ 
        \setlength\tabcolsep{8pt}
        \vspace{5pt}
        \begin{tabular}{@{}lccccc@{}}
            \toprule
            \multirow{2}{*}{\textbf{Method}} &  \multicolumn{3}{c}{\textbf{Image Quality}} & \textbf{Model Size} & \textbf{Frame Rate}\\
            & \textbf{PSNR} $\uparrow$ & \textbf{SSIM} $\uparrow$ & \textbf{LPIPS} $\downarrow$ &\textbf{[MB]} $\downarrow$ & \textbf{[FPS]} $\uparrow$ \\
            \midrule
            \textbf{Ours\textsubscript{full}} & \textbf{25.64} & \textbf{0.845} & \textbf{0.109} & 668 & 72 \\
            \textbf{Ours\textsubscript{w/o init}} & 21.70 & 0.617 & 0.302 & 813 & 45 \\
            \textbf{Ours\textsubscript{w/o depth}} & 24.40 & 0.835 & 0.135 & 671 & 72 \\
            \textbf{Ours\textsubscript{w/o U-Net}} & 24.24 & 0.798 & 0.155 & \textbf{662} & \textbf{78} \\
            \bottomrule
        \end{tabular}
    }
    \caption{%
    Comparison of \textbf{ablated D-NPC versions} in terms of image quality, model size, and speed.%
    }\label{tab:ablation}
\end{table}
To assess the improvements enabled by our main pipeline components, we optimize three variations of our method on the NVIDIA dataset, and compare them to our full D-NPC model (\textit{full}).
First, we train our model from scratch without initializing the probability field from the monocular depth and foreground segmentation, just using camera frustum carving to reduce the amount of initial cells (\textit{w/o init}).
Additionally, we compare against a version without depth supervision during training (\textit{w/o depth}), which is a crucial component in monocular dynamic view synthesis.
Lastly, we evaluate the impact of our neural renderer by skipping the U-Net, directly interpreting the rasterized point cloud features as RGB outputs (\textit{w/o U-Net}).
As evident from \cref{tab:ablation}, our initialization procedure effectively guides the model to a favorable representation, decreasing image quality and increasing memory when omitted.
Likewise, skipping additional depth supervision from the globally aligned monocular depth maps significantly deteriorates reconstruction quality.
The neural rendering network used for hole filling, while also contributing to improved final image quality, comes at the cost of a slightly decreased inference frame rate.
We find that direct RGB rasterization without a rendering network can still produce visually appealing results, and improves the interactivity at higher resolutions when exploring scenes in a graphical user interface.
Importantly, the U-Net weights do not have any significant impact on the final model size, which is dominated by the dynamic sampling and appearance parameters.
\section{Discussion and Limitations}
\label{sec:limitations}
\begin{figure}[bht]
    \centering
    \includegraphics[width=\columnwidth, keepaspectratio]{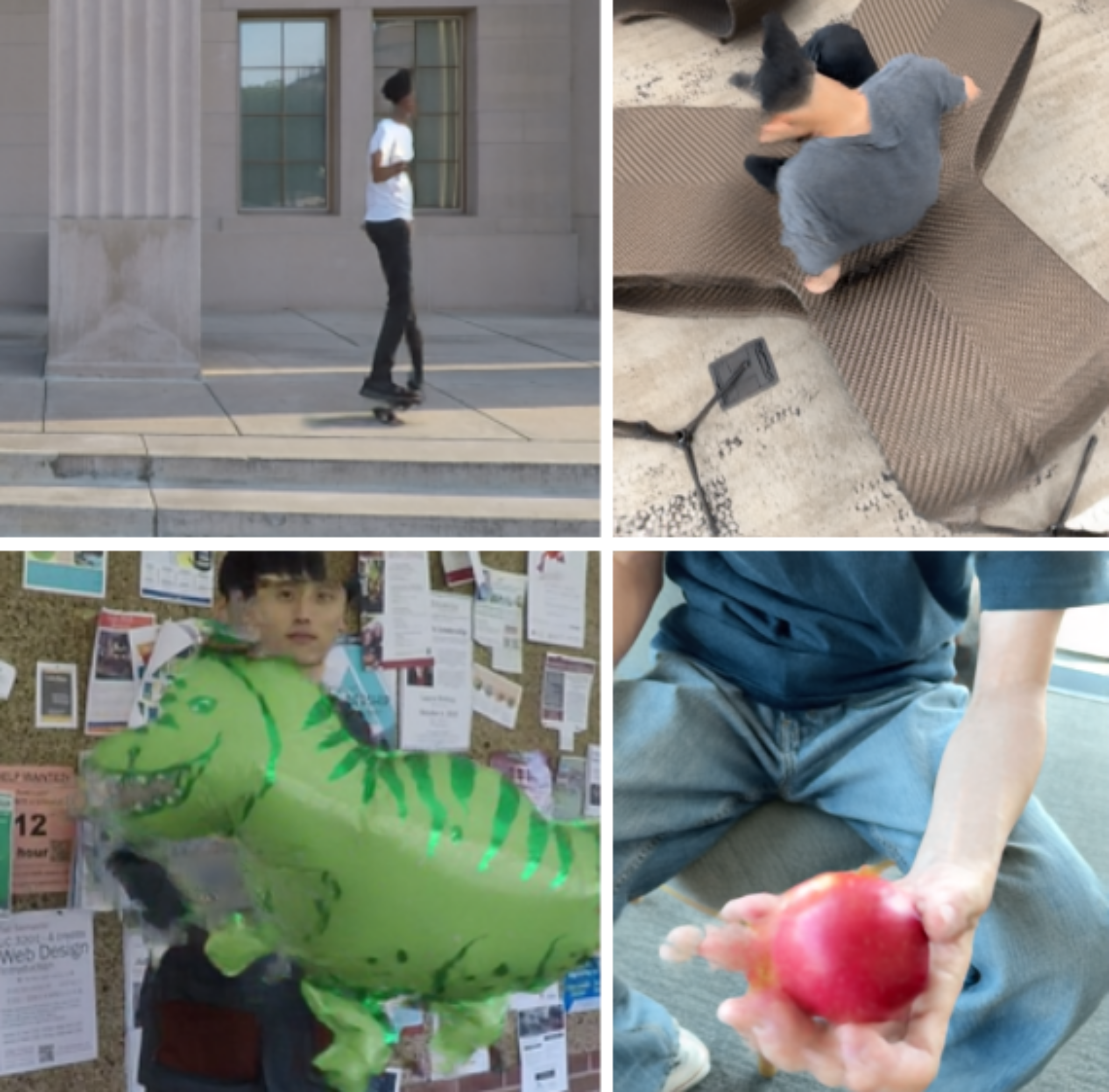} 
    \caption{%
        \textbf{Failure Cases}. Top: Cropped foreground not contained in the training camera frustum. Bottom: Object duplications due to depth misalignment.%
    }\label{fig:limitations} 
\end{figure}

The conducted experiments demonstrate the adaptability and effectiveness of the implicit neural point cloud representation~\cite{hahlbohm2024inpc} for non-rigid novel view synthesis;
Our temporal extension \textit{D-NPC} effectively leverages data-driven priors for dynamic reconstruction from single monocular video, yielding a favorable trade-off between optimization time and image quality.
%
Due to the fast forward-rendering approach, it is currently one of the only purely monocular methods enabling real-time frame rates for interactive scene exploration.
Especially in terms of perceptual LPIPS and high image frequencies, our method quantitatively and visually outperforms most 
recent monocular view synthesis methods while only requiring a fraction of the optimization time.
However, there is room for improvement: Our method relies on a decent quality of camera poses and monocular depth.
Moreover, our initialization process assumes that the dynamic content is fully contained in the camera frustum, resulting in visible crops when synthesizing far-off novel views, as shown in \cref{fig:limitations}.
Finally, our method handles videos of up to $\approx500$ frames (similar to related approaches) using less than $24$GB of GPU-VRAM at FHD resolution, with longer video sequences remaining a limitation.

\section{Conclusion}
\label{sec:conclusion}
We presented \textit{Dynamic Neural Point Clouds}, an efficient implicit point cloud-driven~\cite{hahlbohm2024inpc} approach for non-rigid novel-view synthesis from single monocular video inputs.
In contrast to existing ray marching-based NeRF methods, our rasterization-based forward rendering approach enables a direct and elegant way to initialize and guide the reconstruction from data-driven priors, not only making it one of the fastest approaches in terms of optimization time, but also enabling rendering at real-time frame rates.
While our experiments show our novel approach already yields competitive image quality according to several metrics on common benchmark datasets, we believe that our work constitutes a promising new direction for future research and extensions, like the integration of diffusion-based priors.
%

\section*{Acknowledgments}
We thank Jann-Ole Henningson for help with figures and writing.
This work was partially funded by the DFG under projects “Real-Action VR” (ID 523421583), and “Increasing Realism of Omnidirectional Videos in Virtual Reality” (ID 491805996); the L3S Research Center, Hanover, Germany; as well as by the ERC Consolidator Grant 4DRepLy (770784). Open Access funding enabled and organized by Projekt DEAL.

\bibliographystyle{eg-alpha-doi} 
\bibliography{References-formatted}  

\twocolumn[{ 
\renewcommand\twocolumn[1][]{#1} 
\centering
\vspace{40pt}
\huge{\textbf{D-NPC: Dynamic Neural Point Clouds for \\ Non-Rigid View Synthesis from Monocular Video\\ \vspace{10pt}-- Supplemental Material --}} %
\vspace{40pt}
}] 
%
%
\noindent
In this supplemental, we provide details on our model configuration (\cref{sec:appendix_model}) and optimization procedure (\cref{sec:appendix_optim}), include additional results (\cref{sec:appendix_results}), and discuss the societal impact (\cref{sec:appendix_societal}) of our work.
For dynamic view synthesis results, please see the attached videos.
\section{Model Details}\label{sec:appendix_model}
In our model, the crucial neural components regarding the image quality are the appearance feature grids and the neural rendering network.
To model our static and dynamic appearance fields, we use the fast multi-resolution hash encoding implementation by Müller~\etal~\cite{muller2022instant, tiny-cuda-nn}
After initially carving our sampling distribution voxel grid using the monocular depth and dynamic segmentation priors, we set the spatial scale of our 3D and 4D feature grids to the minimum axis-aligned bounding box enclosing all remaining static and dynamic voxels, respectively.
We parameterize the hash-grids according to the values in \cref{tab:hash-grids}, giving more total capacity to the higher dimensional dynamic representation.
Both hash encodings are followed by a shallow ReLU-activated MLP with one hidden $64$-neuron layer to resolve hash collisions.
For our neural rendering network used for hole-filling, we apply a simple image-to-image translation approach based on a U-Net architecture~\cite{ronneberger2015u}.
In contrast to Hahlbohm~\etal~\cite{hahlbohm2024inpc}, we do not use any multi-resolution inputs.
Instead, we use two downsampling and consecutive upsampling blocks, each consisting of two ReLU-activated 2D convolutions with kernel size $3$, followed by average pooling / bilinear upsampling.
During down- and up-scaling, each convolution block doubles/halves the amount of feature channels while adjusting the spatial resolution, going from an initial 64 channels at the finest to a maximum of 256 channels at the coarsest spatial resolution.

 \begin{table}[bht]
    \centering
    \small{ 
        \begin{tabular}{@{}p{3.4cm} c c}
            \toprule
            \textbf{Parameter} & \textbf{3D Static Grid} & \textbf{4D Dynamic Grid} \\
            \midrule
            Base Resolution & 16 & 16 \\
            Resolution Levels & 10 & 8 \\
            Per Level Scale & 2 & 2 \\
            Features Per Level & 4 & 4 \\
            Logarithmic Size & 21 & 22 \\
            \bottomrule
        \end{tabular}
    }
    \caption{%
        \textbf{Appearance Field Hash Grid Configurations}.%
    }\label{tab:hash-grids}    
\end{table}
\begin{table}[thb]
    \centering
    \small{ 
        \vspace{5pt}
        \begin{tabular}{@{}p{4.5cm} r}
            \toprule
            \textbf{Hyperparameter} & \textbf{Value} \\
            \midrule
            $\lambda_{\text{photo, Cauchy}}$ & 1.00 \\
            $\lambda_{\text{photo, D-SSIM}}$ & 0.90 \\
            $\lambda_{\text{photo, LPIPS}}$ & 0.05 \\
            $\lambda_{\text{depth}}$ & 20.00 \\
            $\lambda_{\text{seg}}$ & $1\mathrm{e}{-3}$ \\
            $\lambda_{\text{dist}}$ & 1.00  \\
            $k$ & 0.50 \\
            \bottomrule
        \end{tabular}
    }
    \caption{%
    \textbf{Loss Hyperparameters}.%
    }\label{tab:hyperparams}
    \vspace{-20pt}
\end{table}
\section{Optimization Details}
\label{sec:appendix_optim}
We train our model for a total of 10K iterations on a consumer-grade RTX 4090 GPU.
During each iteration, we sample an explicit point cloud consisting of $4$M points (\ie a batch of spatial 3D locations).
While the sampling distribution is manually updated based on statistics of the extracted explicit point clouds (see Sec. 3.2.1.3 of the main paper), 
the differentiable rasterizer and neural rendering network enable end-to-end optimization using gradient descent for the network and hash-grid parameters.
We use the Adam~\cite{kingma2014adam} optimizer ($\beta_{1} = 0.9$, $\beta_{2} = 0.99$, $\epsilon = 1\mathrm{e}{-15}$) with an initial learning rate of $1\mathrm{e}{-2}$ for the appearance hash-grids and $5\mathrm{e}{-4}$ for the rendering network, which are exponentially decayed to $\frac{1}{3}\mathrm{e}{-3}$ and $5\mathrm{e}{-5}$ during optimization.
For our loss hyperparameters (Sec. 3.4 of the main paper), we use the values provided in \cref{tab:hyperparams}.
Note that, after initially training with a high value $\lambda_{depth}$ to guide the optimization in a reasonable direction, we turn off our depth loss after $500$ training iterations to allow the model to recover from errors in the monocular depth estimates.
In contrast to D\textsuperscript{2}NeRF, who use $k=2$ to bias the binary segmentation loss towards static background, which usually constitutes a majority of the scene, we use a value of $0.5$ to favor the dynamic model.
This is due to the fact that our initialization procedure already discards large amounts of dynamic model evaluation in static areas.
As a result, the remaining dynamic samples are already biased towards potentially dynamic areas, which we reflect in the choice of skew $k$.
In addition to the losses already discussed in Sec. 3.4 in the main paper, 
we further apply weight decay (\ie an L2 loss) with a factor of $2.5$ on the appearance grid parameters~\cite{barron2023zip}.
In practice, we find that this decay helps to prevent overfitting and improves the quality of generated novel views.
Our Python / CUDA implementation is available online:
{\small\url{https://moritzkappel.github.io/projects/dnpc/}}.
\begin{table*}[tbh]
    \centering
    \scriptsize{ 
        \setlength\tabcolsep{4.5pt}
        \begin{tabular}{@{}lcccccccccccc}
            \toprule
            & \multicolumn{3}{c}{\textbf{Apple}} & \multicolumn{3}{c}{\textbf{Block}} & \multicolumn{3}{c}{\textbf{Paper-windmill}} & \multicolumn{3}{c}{\textbf{Space-out}}\\
            \textbf{Method} & \textbf{mPSNR} $\uparrow$ & \textbf{mSSIM} $\uparrow$ & \textbf{mLPIPS} $\downarrow$ & \textbf{mPSNR} $\uparrow$ & \textbf{mSSIM} $\uparrow$ & \textbf{mLPIPS} $\downarrow$ & \textbf{mPSNR} $\uparrow$ & \textbf{mSSIM} $\uparrow$ & \textbf{mLPIPS} $\downarrow$ & \textbf{mPSNR} $\uparrow$ & \textbf{mSSIM} $\uparrow$ & \textbf{mLPIPS} $\downarrow$\\ 
            \midrule
            HyperNeRF+B+D+S~\cite{park2021hypernerf} & \cellcolor{3rd}17.64& 0.743 & 0.478& \cellcolor{3rd}17.54& \cellcolor{1st}0.670& \cellcolor{2nd}0.331& 17.38 & \cellcolor{3rd}0.382 & \cellcolor{1st}0.209 & \cellcolor{3rd}17.93 & 0.605 & \cellcolor{3rd}0.320 \\
            Nerfies+B+D+S~\cite{park2021nerfies} & 17.54 & \cellcolor{3rd}0.750& 0.478& 16.61 & 0.639 & 0.389 & 17.34 & 0.378 & \cellcolor{2nd}0.211  & 17.79 & \cellcolor{2nd}0.622 & \cellcolor{2nd}0.303 \\
            T-NeRF~\cite{gao2022Neurips} & 16.64 & 0.719 & 0.559 & 16.36 & 0.637 & 0.434 & 15.29 & 0.278 & 0.447  & 17.17 & 0.602 & 0.397 \\
            T-NeRF+B+D~\cite{gao2022Neurips} & 17.32 & 0.720 & 0.514 & \cellcolor{2nd}17.60& \cellcolor{1st}0.670& 0.375 & \cellcolor{2nd}17.56& 0.351 & 0.295  & \cellcolor{3rd}18.34& \cellcolor{3rd}0.619 & 0.345\\
            T-NeRF+B+D+S~\cite{gao2022Neurips} & 17.43 & 0.728 & 0.508 & 17.52 & \cellcolor{2nd}0.669& \cellcolor{3rd}0.346& 17.55 & 0.367 & 0.258  & 17.71 & 0.591 & 0.377\\
            RoDynRF~\cite{liu2023CVPR} & \cellcolor{2nd}18.74& 0.723 & 0.552 & \cellcolor{1st}18.05& 0.634 & 0.513 & 16.71 & 0.321 & 0.482  & \cellcolor{2nd}18.57& 0.594 & 0.413\\
            4D-GS~\cite{wu20234d} & 15.35 & 0.686 & \cellcolor{2nd}0.432& 13.40 & 0.544 & 0.506 & 14.48 & 0.201 & 0.336  & 14.37 & 0.513 & 0.366\\ 
            \midrule
            \textbf{Ours} & 16.83 & \cellcolor{2nd} 0.752 & \cellcolor{3rd}0.469 & 15.53 & 0.632 & 0.350 & \cellcolor{1st}18.01& \cellcolor{1st}0.432& \cellcolor{1st}0.209& \cellcolor{1st}18.69& \cellcolor{1st}0.640& \cellcolor{1st}0.246\\
            \textbf{Ours$_{\text{Lidar}}$} & \cellcolor{1st}19.18& \cellcolor{1st}0.767& \cellcolor{1st}0.393& 16.25 & 0.646 & \cellcolor{1st}0.306& \cellcolor{1st}18.01& \cellcolor{2nd}0.416& 0.218 & 16.25 & 0.614 & 0.345 \\
            \bottomrule
        \end{tabular}
        \begin{tabular}{@{}lccccccccc|ccc}
            & \multicolumn{3}{c}{\textbf{Spin}} & \multicolumn{3}{c}{\textbf{Teddy}} & \multicolumn{3}{c|}{\textbf{Wheel}}& \multicolumn{3}{c}{\textbf{Average}} \\
            \textbf{Method} & \textbf{mPSNR} $\uparrow$ & \textbf{mSSIM} $\uparrow$ & \textbf{mLPIPS} $\downarrow$ & \textbf{mPSNR} $\uparrow$ & \textbf{mSSIM} $\uparrow$ & \textbf{mLPIPS} $\downarrow$ & \textbf{mPSNR} $\uparrow$ & \textbf{mSSIM} $\uparrow$ & \textbf{mLPIPS} $\downarrow$ & \textbf{mPSNR} $\uparrow$ & \textbf{mSSIM} $\uparrow$ & \textbf{mLPIPS} $\downarrow$\\ 
            \midrule
            HyperNeRF+B+D+S~\cite{park2021hypernerf} & \cellcolor{1st}19.20& 0.561 & 0.325 & \cellcolor{3rd}13.97 & \cellcolor{3rd}0.568 & \cellcolor{1st}0.350 & 13.99 & 0.455 & 0.310 & 16.81 & 0.569 & \cellcolor{2nd}0.332 \\
            Nerfies+B+D+S~\cite{park2021nerfies} & 18.38 & \cellcolor{1st}0.585& \cellcolor{2nd}0.309 & 13.65 & 0.557 & \cellcolor{2nd}0.372 & 13.82 & 0.458 & 0.310 & 16.45 & 0.570 & \cellcolor{3rd}0.339 \\
            T-NeRF~\cite{gao2022Neurips} & 16.36 & 0.504 & 0.530 & 13.38 & 0.550 & 0.475 & 14.73 & 0.477 & 0.362 & 15.70 & 0.538 & 0.458 \\
            T-NeRF+B+D~\cite{gao2022Neurips} & \cellcolor{3rd}18.95& 0.541 & 0.468 & \cellcolor{2nd}13.99 & \cellcolor{1st}0.573 & 0.435 & \cellcolor{1st}15.69& 0.541 & \cellcolor{3rd}0.301 & \cellcolor{1st}17.06 & 0.573 & 0.390 \\
            T-NeRF+B+D+S~\cite{gao2022Neurips} & \cellcolor{2nd}19.16& 0.567& 0.443 & 13.71 & \cellcolor{2nd}0.570 & 0.429 & \cellcolor{2nd}15.65& \cellcolor{3rd}0.548 & \cellcolor{2nd}0.292 & \cellcolor{3rd}16.96 & \cellcolor{3rd}0.577 & 0.379 \\
            RoDynRF~\cite{liu2023CVPR} & 17.41 & 0.484 & 0.570 & \cellcolor{1st}14.34 & 0.537 & 0.613 & 15.20 & 0.449 & 0.478 & \cellcolor{2nd}17.00 & 0.535 & 0.517 \\
            4D-GS~\cite{wu20234d} & 15.36 & 0.418 & 0.310& 12.36 & 0.506 & 0.474 & 11.78 & 0.343 & 0.427 & 13.87 & 0.459 & 0.407 \\ 
            \midrule
            \textbf{Ours} & 17.78 & \cellcolor{1st}0.585& \cellcolor{2nd}0.309& 12.19 & 0.536 & 0.503 & 13.27 & \cellcolor{2nd}0.549& 0.349 & 16.04 & \cellcolor{2nd}0.589& 0.348\\
            \textbf{Ours$_{\text{Lidar}}$} & 18.55 & \cellcolor{2nd}0.573& \cellcolor{1st}0.296& 13.63 & 0.557 & \cellcolor{3rd}0.418& \cellcolor{3rd}15.63& \cellcolor{1st}0.609& \cellcolor{1st}0.274& 16.79 & \cellcolor{1st}0.597& \cellcolor{1st}0.321\\
            \bottomrule
        \end{tabular}
 %
    }
    \caption{%
    \textbf{Quantitative evaluation on the iPhone dataset}. We compare our method against related approaches using masked metrics based on co-visibility~\cite{gao2022Neurips}. Methods marked with $\dagger$ are trained using the code and additional regularizers introduced by Gao~\etal~\cite{gao2022Neurips}. We further show values for our method trained on monocular depth instead of the provided Lidar data. The three best results are highlighted in \textcolor{1stText}{\textbf{green}} in descending order of saturation.%
    }\label{tab:iphone}
\end{table*}
\begin{table*}[thb]
    \centering
    \scriptsize{ 
    \setlength\tabcolsep{7pt}
    \begin{tabular}{@{}lcccccccccccc}
    \toprule
  & \multicolumn{3}{c}{\textbf{Balloon1}} & \multicolumn{3}{c}{\textbf{Balloon2}} & \multicolumn{3}{c}{\textbf{Jumping}} & \multicolumn{3}{c}{\textbf{Playground}}\\
    \textbf{Method} & \textbf{PSNR} $\uparrow$ & \textbf{SSIM} $\uparrow$ & \textbf{LPIPS} $\downarrow$ & \textbf{PSNR} $\uparrow$ & \textbf{SSIM} $\uparrow$ & \textbf{LPIPS} $\downarrow$ & \textbf{PSNR} $\uparrow$ & \textbf{SSIM} $\uparrow$ & \textbf{LPIPS} $\downarrow$ & \textbf{PSNR} $\uparrow$ & \textbf{SSIM} $\uparrow$ & \textbf{LPIPS} $\downarrow$\\ 
    \midrule
    T-NeRF~\cite{pumarola2021d} & 18.54 & 0.448 & 0.416 & 20.69 & 0.573 & 0.335 & 18.04 & 0.496 & 0.541 & 14.68 & 0.240 & 0.527\\
    D-NeRF~\cite{pumarola2021d} & 19.06 & 0.492 & 0.382 & 20.76 & 0.557 & 0.367 & 22.36 & 0.712 & 0.318 & 20.18 & 0.670 & 0.250\\
    HyperNeRF~\cite{park2021hypernerf} & 13.96 & 0.214 & 0.677 & 16.57 & 0.286 & 0.544 & 18.34 & 0.521 & 0.460 & 13.17 & 0.179 & 0.603\\
    NR-NeRF~\cite{tretschk2021non} & 17.39 & 0.369 & 0.497 & 22.41 & 0.703 & 0.314 & 20.09 & 0.644 & 0.418 & 15.06 & 0.248 & 0.450\\
    NSFF~\cite{li2021neural} & 21.96 & 0.701& 0.281 & 24.27 & 0.731 & 0.276 & \cellcolor{3rd}24.65& 0.813 & 0.227 & 21.22 & 0.705 & 0.263\\
    DynamicNeRF~\cite{gao2021dynamic} & \cellcolor{2nd}22.36& \cellcolor{3rd}0.775& \cellcolor{3rd}0.169& \cellcolor{1st}27.06& \cellcolor{1st}0.859& \cellcolor{2nd}0.113& \cellcolor{2nd}24.68& \cellcolor{2nd}0.842& \cellcolor{3rd}0.153& \cellcolor{2nd}24.15& \cellcolor{3rd}0.849& \cellcolor{3rd}0.145\\
    RoDynRF~\cite{liu2023CVPR} & \cellcolor{1st}22.37& \cellcolor{1st}0.782& \cellcolor{1st}0.157& \cellcolor{2nd}26.19& \cellcolor{2nd}0.846& \cellcolor{1st}0.107& \cellcolor{1st}25.66& \cellcolor{1st}0.853& \cellcolor{2nd}0.129& \cellcolor{1st}24.96& \cellcolor{1st}0.899& \cellcolor{1st}0.080\\
    TiNeuVox~\cite{fang2022fast} & 17.30 & 0.392 & 0.500 & 19.06 & 0.458 & 0.396 & 20.81 & 0.641 & 0.399 & 13.84 & 0.213 & 0.541\\
    4D-GS~\cite{wu20234d} & 20.82 & 0.666 & 0.271 & 24.34 & 0.775 & 0.211 & 22.16 & 0.735 & 0.270 & 20.22 & 0.724 & 0.192\\
    \midrule
    \textbf{Ours} & \cellcolor{3rd}22.26& \cellcolor{2nd}0.779& \cellcolor{2nd}0.161& \cellcolor{3rd}25.84& \cellcolor{3rd}0.835& \cellcolor{1st}0.107& 24.51 & \cellcolor{3rd}0.839& \cellcolor{1st}0.116&  \cellcolor{3rd}23.59& \cellcolor{2nd}0.866&  \cellcolor{2nd}0.099\\ 
    \bottomrule
    \end{tabular}
     \begin{tabular}{@{}lccccccccc|ccc}
    & \multicolumn{3}{c}{\textbf{Skating}} & \multicolumn{3}{c}{\textbf{Truck}} & \multicolumn{3}{c|}{\textbf{Umbrella}}& \multicolumn{3}{c}{\textbf{Average}} \\
    \textbf{Method} & \textbf{PSNR} $\uparrow$ & \textbf{SSIM} $\uparrow$ & \textbf{LPIPS} $\downarrow$ & \textbf{PSNR} $\uparrow$ & \textbf{SSIM} $\uparrow$ & \textbf{LPIPS} $\downarrow$ & \textbf{PSNR} $\uparrow$ & \textbf{SSIM} $\uparrow$ & \textbf{LPIPS} $\downarrow$ & \textbf{PSNR} $\uparrow$ & \textbf{SSIM} $\uparrow$ & \textbf{LPIPS} $\downarrow$\\ 
    \midrule
    T-NeRF~\cite{pumarola2021d} & 20.32 & 0.551 & 0.562 & 18.33 & 0.462 & 0.523 & 17.69 & 0.284 & 0.670 & 18.33 & 0.436 & 0.511\\
    D-NeRF~\cite{pumarola2021d} & 22.48 & 0.678 & 0.403 & 24.10 & 0.743 & 0.213 & 21.47 & 0.554 & 0.310 & 21.49 & 0.629 & 0.320\\
    HyperNeRF~\cite{park2021hypernerf} & 21.97 & 0.640 & 0.322 & 20.61 & 0.493 & 0.394 & 18.59 & 0.308 & 0.517 & 17.60 & 0.377 & 0.502\\
    NR-NeRF~\cite{tretschk2021non} & 23.95 & 0.745 & 0.342 & 19.33 & 0.485 & 0.548 & 19.63 & 0.376 & 0.471 & 19.69 & 0.510 & 0.434\\
    NSFF~\cite{li2021neural} & \cellcolor{3rd}29.29& 0.889 & 0.186 & 25.96 & 0.775 & 0.248 & 22.97 & 0.644 & 0.315 & 24.33 & 0.751 & 0.257\\
    DynamicNeRF~\cite{gao2021dynamic} & \cellcolor{1st}32.66& \cellcolor{1st}0.951& \cellcolor{3rd}0.077& \cellcolor{3rd}28.56& \cellcolor{3rd}0.872& \cellcolor{3rd}0.146& \cellcolor{3rd}23.26& \cellcolor{3rd}0.711 & \cellcolor{3rd}0.190& \cellcolor{1st}26.10& \cellcolor{3rd}0.837& \cellcolor{3rd}0.142\\
    RoDynRF~\cite{liu2023CVPR} & 28.68 & \cellcolor{3rd}0.939& \cellcolor{2nd}0.076& \cellcolor{1st}{29.13}& \cellcolor{1st}{0.900}& \cellcolor{2nd}0.089& \cellcolor{1st}{24.26}& \cellcolor{1st}0.757 & \cellcolor{1st}{0.135}& \cellcolor{2nd}25.89& \cellcolor{1st}{0.854}& \cellcolor{2nd}0.110\\
    TiNeuVox~\cite{fang2022fast} & 23.32 & 0.723 & 0.283 & 23.86 & 0.662 & 0.328 & 20.00 & 0.416 & 0.439 & 19.74 & 0.501 & 0.412\\
    4D-GS~\cite{wu20234d} & 26.04 & 0.876 & 0.171 & 25.08 & 0.753 & 0.233 & 21.59 & 0.588 & 0.309 & 22.89 & 0.731 & 0.237\\
    \midrule
    \textbf{Ours} & \cellcolor{2nd}30.22& \cellcolor{2nd}0.948& \cellcolor{1st}0.061& \cellcolor{2nd}28.92& \cellcolor{2nd}0.897& \cellcolor{1st}0.084& \cellcolor{2nd}24.15& \cellcolor{2nd}0.747 & \cellcolor{2nd}0.136& \cellcolor{3rd}25.64& \cellcolor{2nd}0.845& \cellcolor{1st}0.109\\
    \bottomrule
    \end{tabular}
 %
    }
     \caption{
    \textbf{Quantitative comparison on the NVIDIA dataset}. We compare our method against recent dynamic view synthesis approaches on the NVIDIA dynamic scenes dataset. Our method achieves the best scores in terms of perceptual image quality. The three best results are highlighted in \textcolor{1stText}{\textbf{green}} in descending order of saturation.}
    \label{tab:nvidia_short}
\end{table*}
\section{Additional Results}\label{sec:appendix_results}
\begin{table*}[thb]
    \centering
    \scriptsize{ 
        \setlength\tabcolsep{5pt}
        \begin{tabular}{@{}lcccccccccccc}
            \toprule
            & \multicolumn{3}{c}{\textbf{Balloon1}} & \multicolumn{3}{c}{\textbf{Balloon2}} & \multicolumn{3}{c}{\textbf{Jumping}} & \multicolumn{3}{c}{\textbf{Playground}}\\
            \textbf{Method} & \textbf{mPSNR} $\uparrow$ & \textbf{mSSIM} $\uparrow$ & \textbf{mLPIPS} $\downarrow$ & \textbf{mPSNR} $\uparrow$ & \textbf{mSSIM} $\uparrow$ & \textbf{mLPIPS} $\downarrow$ & \textbf{mPSNR} $\uparrow$ & \textbf{mSSIM} $\uparrow$ & \textbf{mLPIPS} $\downarrow$ & \textbf{mPSNR} $\uparrow$ & \textbf{mSSIM} $\uparrow$ & \textbf{mLPIPS} $\downarrow$\\ 
            \midrule
            T-NeRF~\cite{pumarola2021d} & 17.90 & 0.897 & 0.245 & 19.89 & 0.953 & 0.133 & 15.24 & 0.914 & 0.285 & 14.81 & 0.957 & 0.267\\ 
            D-NeRF~\cite{pumarola2021d} & 18.75 & 0.907 & 0.294 & 18.23 & 0.950 & 0.203 & 19.23 & 0.941 & 0.160 & 16.65 & 0.970 & 0.227\\ 
            HyperNeRF~\cite{park2021hypernerf} & 14.43 & 0.865 & 0.471 & 15.95 & 0.931 & 0.251 & 14.82 & 0.913 & 0.278 & 12.32 & 0.950 & 0.390\\
            NR-NeRF~\cite{tretschk2021non}& 15.87 & 0.880 & 0.383 & 20.13 & 0.962 & 0.177 & 16.76 & 0.934 & 0.228 & 13.91 & 0.956 & 0.311\\
            NSFF~\cite{li2021neural} & 19.34 & 0.925 & 0.225 & \cellcolor{3rd}22.63& 0.969 & 0.155 & 20.42 & 0.955 & 0.152 & \cellcolor{3rd}20.22& 0.983 & 0.162\\
            DynamicNeRF~\cite{gao2021dynamic} & \cellcolor{2nd}19.73& \cellcolor{1st}0.931& \cellcolor{1st}0.134& \cellcolor{1st}23.79& \cellcolor{1st}0.981& \cellcolor{1st}0.042& \cellcolor{3rd}20.68& \cellcolor{3rd}0.958& \cellcolor{3rd}0.100& \cellcolor{1st}21.05& \cellcolor{1st}0.984& \cellcolor{1st}0.095\\
            RoDynRF~\cite{liu2023CVPR} & \cellcolor{3rd}19.70& \cellcolor{3rd}0.928& \cellcolor{2nd}0.147& \cellcolor{2nd}23.05& \cellcolor{2nd}0.977& \cellcolor{2nd}0.046& \cellcolor{1st}21.54& \cellcolor{1st}0.962& \cellcolor{2nd}0.093& \cellcolor{2nd}20.29& \cellcolor{1st}0.984& \cellcolor{2nd}0.097\\ 
            TiNeuVox~\cite{fang2022fast} & 16.80 & 0.888 & 0.330 & 19.59 & 0.955 & 0.147 & 17.34 & 0.932 & 0.196 & 13.02 & 0.950 & 0.363\\
            4D-GS~\cite{wu20234d} & 18.72 & 0.911 & 0.247 & 22.21 & \cellcolor{3rd}0.973& 0.066 & 18.60 & 0.938 & 0.173 & 17.23 & 0.976 & 0.156\\
            \midrule
            \textbf{Ours} & \cellcolor{1st}19.76& \cellcolor{2nd}0.930& \cellcolor{3rd}0.159&  22.14 & \cellcolor{3rd}0.973& \cellcolor{3rd}0.056&  \cellcolor{2nd}21.07& \cellcolor{2nd}0.960& \cellcolor{1st}0.080& 19.76 &  \cellcolor{2nd}0.982& \cellcolor{3rd}0.107\\
            \bottomrule
        \end{tabular}
        \begin{tabular}{@{}lccccccccc|ccc}
            & \multicolumn{3}{c}{\textbf{Skating}} & \multicolumn{3}{c}{\textbf{Truck}} & \multicolumn{3}{c|}{\textbf{Umbrella}}& \multicolumn{3}{c}{\textbf{Average}} \\
            \textbf{Method} & \textbf{mPSNR} $\uparrow$ & \textbf{mSSIM} $\uparrow$ & \textbf{mLPIPS} $\downarrow$ & \textbf{mPSNR} $\uparrow$ & \textbf{mSSIM} $\uparrow$ & \textbf{mLPIPS} $\downarrow$ & \textbf{mPSNR} $\uparrow$ & \textbf{mSSIM} $\uparrow$ & \textbf{mLPIPS} $\downarrow$ & \textbf{mPSNR} $\uparrow$ & \textbf{mSSIM} $\uparrow$ & \textbf{mLPIPS} $\downarrow$\\ 
            \midrule
            T-NeRF~\cite{pumarola2021d} & 22.56 & 0.993 & 0.086 & 18.40 & 0.913 & 0.409 & 16.32 & 0.945 & 0.496 & 17.87 & 0.939 & 0.274\\
            D-NeRF~\cite{pumarola2021d}  & 22.05 & 0.993 & 0.090 & 19.84 & 0.923 & 0.312 & 17.34 & 0.953 & 0.365 & 18.87 & 0.948 & 0.236\\
            HyperNeRF~\cite{park2021hypernerf}  & 23.86 & 0.993 & 0.074 & 19.40 & 0.918 & 0.281 & 14.76 & 0.940 & 0.489 & 16.51 & 0.930 & 0.319\\
            NR-NeRF~\cite{tretschk2021non} & 27.17 & 0.995 & 0.052 & 18.25 & 0.912 & 0.534 & 16.23 & 0.948 & 0.404 & 18.33 & 0.941 & 0.298\\
            NSFF~\cite{li2021neural} & 31.37 & \cellcolor{2nd}0.998& \cellcolor{3rd}0.019& 26.51 & 0.971 & 0.135 & 19.28 & 0.967 & 0.238 & 22.82 & 0.967 & 0.155\\
            DynamicNeRF~\cite{gao2021dynamic} & \cellcolor{1st}37.22 & \cellcolor{1st}0.999& \cellcolor{1st}0.010& \cellcolor{3rd}26.75& \cellcolor{3rd}0.973& \cellcolor{3rd}0.067& \cellcolor{3rd}19.66& \cellcolor{3rd}0.969& \cellcolor{3rd}0.152& \cellcolor{1st}24.13& \cellcolor{1st}0.971& \cellcolor{2nd}0.086\\
            RoDynRF~\cite{liu2023CVPR} & \cellcolor{3rd}34.37 & 0.997 & 0.035 & \cellcolor{1st}27.29& \cellcolor{1st}0.978& \cellcolor{1st}0.061& \cellcolor{2nd}20.85& \cellcolor{2nd}0.970& \cellcolor{2nd}0.134& \cellcolor{3rd}23.87& \cellcolor{1st}0.971& \cellcolor{3rd}0.087\\
            TiNeuVox~\cite{fang2022fast}  & 27.18 & 0.996 & 0.047 & 21.39 & 0.934 & 0.204 & 16.09 & 0.949 & 0.358 & 18.77 & 0.943 & 0.235\\
            4D-GS~\cite{wu20234d} & 30.99 & 0.997 & 0.048 & 22.77 & 0.947 & 0.207 & 18.36 & 0.956 & 0.259 & 21.27 & 0.957 & 0.165\\
            \midrule
            \textbf{Ours} & \cellcolor{2nd}36.38 & \cellcolor{1st}0.999& \cellcolor{2nd}0.014& \cellcolor{2nd}27.16& \cellcolor{2nd}0.975& \cellcolor{2nd}0.063& \cellcolor{1st}21.71& \cellcolor{1st}0.975& \cellcolor{1st}0.087& \cellcolor{2nd}24.00& \cellcolor{1st}0.971& \cellcolor{1st}0.081\\
            \bottomrule
        \end{tabular}
    }
    \caption{%
    \textbf{Foreground analysis on the short NVIDIA dataset}. We show additional quantitative analysis on the short NVIDIA sequences. We use masked metrics~\cite{gao2022Neurips} with the provided foreground masks to evaluate image quality on only the dynamic part of the scenes. The three best results are highlighted in \textcolor{1stText}{\textbf{green}} in descending order of saturation.
    }\label{tab:nvidia_short_masked}
\end{table*}

We provide detailed per-scene breakdowns on the iPhone and NVIDIA datasets in \cref{tab:iphone} and \cref{tab:nvidia_short}, respectively.
Our method consistently achieves competitive results across a variety of challenging scenes, and particularly excels at perceptual LPIPS scores due to its ability to retain high-frequency details.
As most natural scenes are dominated by static background, we further assess the individual quality of dynamic foreground reconstruction, which is the most complex part of dynamic view synthesis from monocular inputs.
To this end, we recalculate metrics on the NVIDIA sequences using masked metrics~\cite{gao2022Neurips}, based on the motion masks contained in the dataset.
As shown in \cref{tab:nvidia_short_masked}, our method still performs favorably regarding perceptual metrics, indicating that improvements are not only achieved in the static background, but also in the more challenging dynamic foreground represented in only a single image.
\begin{table*}[thb]
    \centering
    \scriptsize{ 
        \setlength\tabcolsep{10pt}
        \begin{tabular}{@{}llcccccccc|c}
            \toprule
            \textbf{Method}   & \textbf{Metric}   &   \textbf{Balloon1}  &  \textbf{Balloon2}  &  \textbf{Jumping}  &  \textbf{Playground}  &  \textbf{Skating}  &  \textbf{Truck}  &  \textbf{Umbrella}  &  \textbf{DynamicFace}  & \textbf{Average}  \\
            \midrule
            DynIBaR~\cite{li2023CVPR}  &   \multirow{2}{*}{\textbf{PSNR} $\uparrow$}&  28.99    &   28.90    &   23.74   &    28.32     &   31.50   &  33.75  &   28.25    &     29.18     & 29.08  \\
            \textbf{Ours} &&   27.32    &   30.09    &   22.67   &    25.51     &   33.29   &  32.14  &   24.76    &     28.18     & 28.00  \\
            \midrule
             DynIBaR~\cite{li2023CVPR}  &   \multirow{2}{*}{\textbf{SSIM} $\uparrow$}&  0.920    &   0.942    &   0.866   &    0.953     &   0.969   &  0.966  &   0.898    &     0.971     & 0.936  \\
             \textbf{Ours} &&   0.902    &   0.934    &   0.841   &    0.895     &   0.969   &  0.952  &   0.737    &     0.962     & 0.899  \\
            \midrule
            DynIBaR~\cite{li2023CVPR}  &   \multirow{2}{*}{\textbf{LPIPS} $\downarrow$} &   0.064    &   0.070    &   0.126   &    0.049     &   0.058   &  0.049  &   0.085    &     0.052     & 0.069  \\
            \textbf{Ours} &&   0.065    &   0.052    &   0.128   &    0.072     &   0.044   &  0.045  &   0.142    &     0.048     & 0.074  \\
            \midrule
            DynIBaR~\cite{li2023CVPR}  &   \multirow{2}{*}{\textbf{mPSNR} $\uparrow$} &     23.91    &   25.53    &   17.49   &    20.91     &   19.52   &  29.06  &   25.96    &     30.55     & 24.12  \\
            \textbf{Ours} &&   22.35    &   23.51    &   16.30   &    18.30     &   19.89   &  25.57  &   21.75    &     23.94     & 21.45  \\
            \midrule
            DynIBaR~\cite{li2023CVPR}  &    \multirow{2}{*}{\textbf{mSSIM} $\uparrow$}&   0.944    &   0.985    &   0.911   &    0.989     &   0.990   &  0.988  &   0.971    &     0.997     & 0.972  \\
            \textbf{Ours} &&   0.937    &   0.980    &   0.896   &    0.984     &   0.992   &  0.979  &   0.921    &     0.991     & 0.960  \\
            \midrule
            DynIBaR~\cite{li2023CVPR}  &   \multirow{2}{*}{\textbf{mLPIPS} $\downarrow$} &  0.079    &   0.032    &   0.168   &    0.062     &   0.070   &  0.038  &   0.043    &     0.010     & 0.062  \\
            \textbf{Ours} &&   0.077    &   0.037    &   0.193   &    0.097     &   0.069   &  0.053  &   0.110    &     0.026     & 0.083  \\
            \bottomrule
        \end{tabular}
    }
    \caption{%
        \textbf{Comparison on the long NVIDIA dataset}. We compare to the state-of-the-art approach DynIBaR~\cite{li2023CVPR} on the long sequences from the NVIDIA dynamic scenes dataset. Masked metrics are calculated with respect to the dynamic foreground masks.%
    }\label{tab:nvidia_long}
\end{table*}
Given that our method can handle significantly longer and temporally denser sequences than the commonly used train/test splits of the NVIDIA dynamic scenes dataset, we further compare against DynIBaR~\cite{li2023CVPR}, using their evaluation protocol containing $\sim$150 training views per scene.
For fairness, we also use the consistent video depth made available by the authors.
Results are shown in ~\cref{tab:nvidia_long}.
While our method attains an overall lower yet still competitive image quality, it requires significantly lower training and rendering times, making it a feasible alternative for real-time applications.
\begin{figure*}[ht]
    \centering
    \vspace{-4pt}
    \setlength\tabcolsep{0pt}
    \small{
        \begin{tabular}{*{6}{p{0.15832\linewidth}<{\centering}}}
            Ground Truth & 4D-GS & Ours & Ground Truth & 4D-GS & Ours
        \end{tabular}
    }
    \includegraphics[width=.94\linewidth, keepaspectratio]{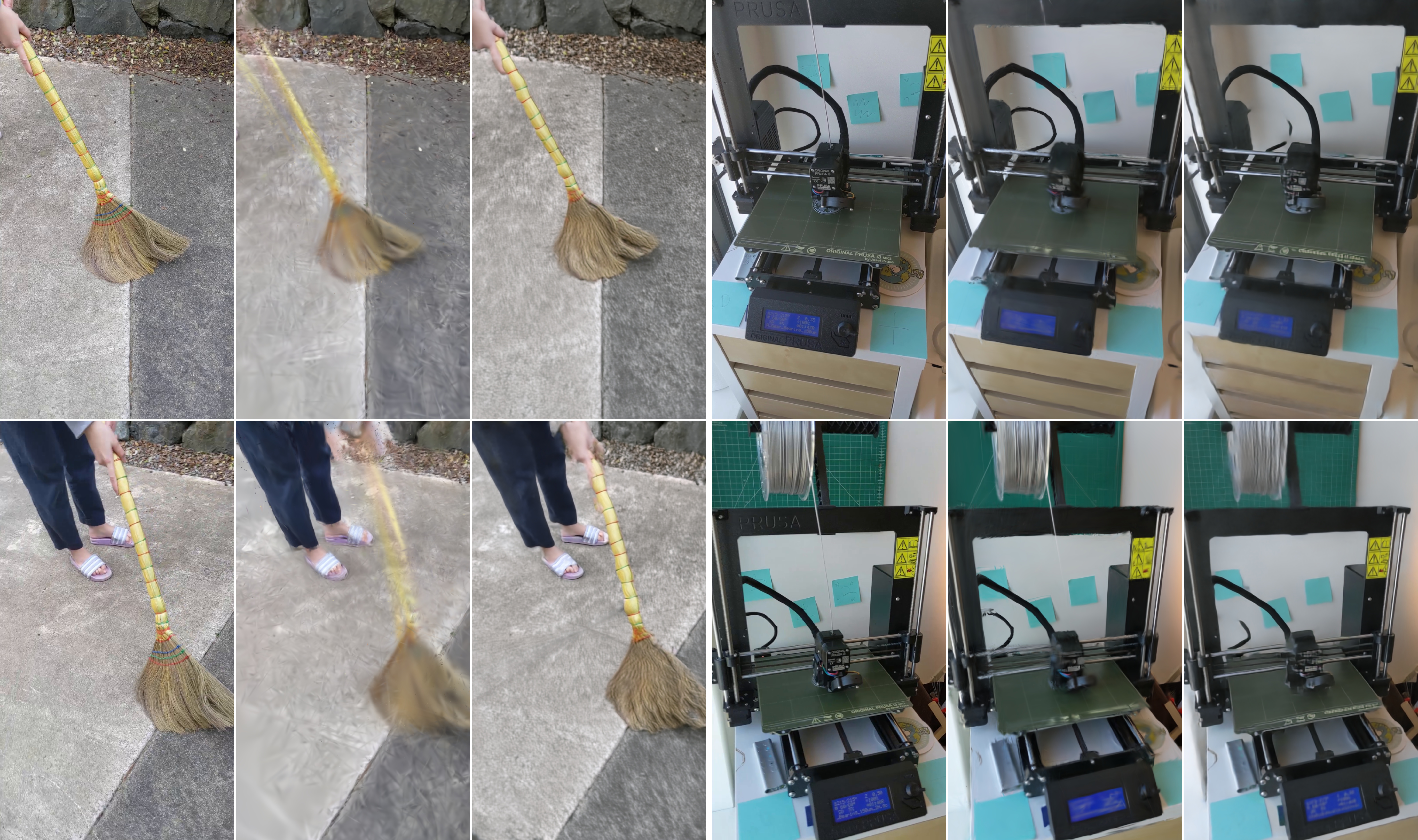}
    \vspace{-8pt}
    \caption{%
        \textbf{Visual comparison with 4D-GS on two scenes from the HyperNeRF dataset.}\vspace{-8pt}%
    }\label{fig:hypernerf}
\end{figure*}
\begin{table*}[thb]
    \centering
    \vspace{-5pt}
    \tiny{
        \setlength\tabcolsep{4.2pt}
        \begin{tabular}{@{}lcccccccccccc|ccc}
            \toprule
            & \multicolumn{3}{c}{\textbf{3D Printer}}                         & \multicolumn{3}{c}{\textbf{Broom}}                              & \multicolumn{3}{c}{\textbf{Peel Banana}}                        & \multicolumn{3}{c|}{\textbf{Chicken}}                            & \multicolumn{3}{c}{\textbf{Average}}                               \\
            \textbf{Method} & \textbf{LPIPS} $\downarrow$ & \textbf{SSIM} $\uparrow$ & \textbf{PSNR} $\uparrow$ & \textbf{LPIPS} $\downarrow$ & \textbf{SSIM} $\uparrow$ & \textbf{PSNR} $\uparrow$ & \textbf{LPIPS} $\downarrow$ & \textbf{SSIM} $\uparrow$ & \textbf{PSNR} $\uparrow$ & \textbf{LPIPS} $\downarrow$ & \textbf{SSIM} $\uparrow$ & \textbf{PSNR} $\uparrow$ & \textbf{LPIPS} $\downarrow$ & \textbf{SSIM} $\uparrow$ & \textbf{PSNR} $\uparrow$ \\ 
            \midrule
            4D-GS~\cite{wu20234d} & 0.367              & 0.701           & 22.03           & 0.624              & 0.352           & 21.88           & 0.231              & 0.853           & 28.03           & 0.316              & 0.812           & 28.74           & 0.384              & 0.680           & 25.17           \\
            \textbf{Ours}  & 0.316              & 0.713           & 21.81           & 0.518              & 0.359           & 21.11           & 0.292              & 0.753           & 23.78           & 0.235              & 0.821           & 25.86           & 0.340              & 0.661           & 23.14           \\ 
            \bottomrule
        \end{tabular}
    }
    \vspace{-4pt}
    \caption{%
        \textbf{Quantitative comparison with 4D-GS on the HyperNeRF dataset.}%
    }\label{tab:hypernerf}
\end{table*}
\begin{figure*}[!ht]
    \centering
    \vspace{-6pt}
    \setlength\tabcolsep{0pt}
    \small{
        \begin{tabular}{*{6}{p{0.15832\linewidth}<{\centering}}}
            Ground Truth & PGDVS & Ours & Ground Truth & PGDVS & Ours
        \end{tabular}
    }
    \includegraphics[width=.94\linewidth, keepaspectratio]{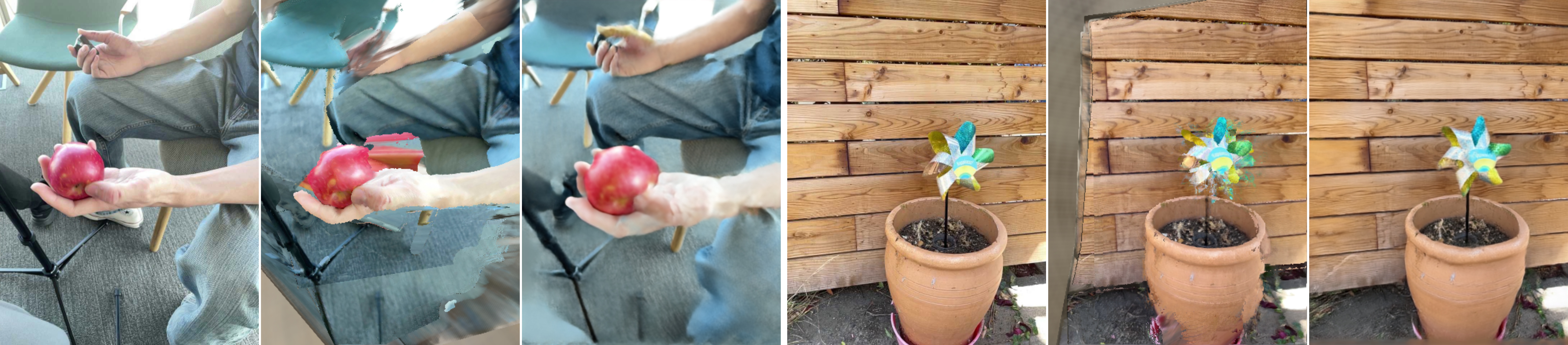}
    \vspace{-8pt}
    \caption{%
        Visual comparison with PGDVS on two scenes from the iPhone dataset.
    }\label{fig:pgdvs}
\end{figure*}
\begin{table*}[!ht]
\centering
\vspace{-4pt}
\scriptsize{
\setlength\tabcolsep{9pt}
\begin{tabular}{@{}lcccc|ccc|cc}
\toprule
             & \multicolumn{2}{c}{\textbf{Paper Windmill}} & \multicolumn{2}{c|}{\textbf{Apple}}        & \multicolumn{3}{c|}{\textbf{Average}}         & \multicolumn{2}{c}{\textbf{Computational Cost}}                    \\
             \textbf{Method} & \textbf{mLPIPS} $\downarrow$  & \textbf{mPSNR} $\uparrow$ & \textbf{mLPIPS} $\downarrow$ & \textbf{mPSNR} $\uparrow$ & \textbf{mLPIPS} $\downarrow$ & \textbf{mPSNR} $\uparrow$ & \textbf{mSSIM} $\uparrow$  & \textbf{Training [hrs]} $\downarrow$ & \textbf{Frame Rate [FPS]} $\uparrow$ \\ \midrule
PGDVS~\cite{Zhao2024PGDVS}        & 0.277                & 17.19            & 0.411               & 16.66            & 0.340               & 15.88            & 0.548            & 0                           & 0.01                        \\
\textbf{Ours }        & 0.209                & 18.03            & 0.487               & 16.45            & 0.354               & 15.79            & 0.585                 & 0.29                        & 65                          \\
\textbf{Ours$_{\text{Lidar}}$ }& 0.212                & 17.99            & 0.414               & 18.79            & 0.319               & 16.84            & 0.597                 & 0.28                        & 65                          \\ \bottomrule
\end{tabular}
}
\vspace{-2pt}
\caption{
\textbf{Quantitative comparison with PGDVS on the iPhone dataset.} We use the official implementation to generate "Paper Windmill" and "Apple" results. Average values correspond to the full dataset and for PGDVS they are copied from the paper. Computational cost was measured using an RTX 4090. To measure FPS for PGDVS we follow the instructions provided with the implementation. Note that, due to the low inference performance achieved by PGDVS, rendering a short five second video at $30$ FPS would take more than $3$ hours. Our method achieves the same in less than $30$ minutes ($\sim17$min training, $\sim3$s rendering).
}\label{tab:pgdvs}
\end{table*}
We further include qualitative and quantitative comparisons to 4D-GS~\cite{wu20234d} on  prominent scenes from the HyperNeRF dataset~\cite{park2021hypernerf} in \cref{fig:hypernerf} and \cref{tab:hypernerf} respectively.
Despite the monocularized setup, where training images are sampled from a stereo camera setup in an alternating manner, we observe the same overall behavior as for real monocular inputs that our method was designed for.
Specifically, our method produces more articulated novel views with occasional slight depth misalignments, resulting in better perceptual scores (SSIM, LPIPS), but worse average PSNR than the blurrier 4D-GS outputs.
We also show comparisons against the generalized PGDVS~\cite{Zhao2024PGDVS} approach on the iPhone dataset.
Despite the theoretical advantage of not requiring per-scene optimization, PGDVS renders novel views at a significantly slower rate, resulting in a longer total inference time on the test set ($\sim$400 images) than the combined training and inference times of our method.
What is more, our per scene optimization naturally enables view synthesis at a higher quality, as shown in \cref{fig:pgdvs} and \cref{tab:pgdvs}.

 \begin{table}[bth]
    \centering
    \vspace{12pt}
    \small{ 
    \begin{tabular}{@{}p{4.5cm} r}
        \toprule
        \textbf{Component} & \textbf{Time [s]}\\
        \midrule
         Data Preprocessing & 140\\
        Depth Alignment & 6\\
        Model Initialization & 1\\
        Model Optimization & 817\\
        \bottomrule
    \end{tabular}
    }
     \caption{
    \textbf{Timings for individual method components}.}
    \label{tab:pipeline-timings}
\end{table}
Finally, we provide a breakdown of timings for individual components of our method on the NVIDIA dataset in \cref{tab:pipeline-timings}.
Note that the cost of data preprocessing highly depends on the amount of input images, while the optimization and rendering times increase with image resolution.
%
\section{Societal Impact}
\label{sec:appendix_societal}
Our method reconstructs a 4D representation and generates novel views based on a monocular input video.
As the method itself cannot fill unseen areas or use/provide any information that goes beyond the inputs provided by the users themselves, we do not expect any negative societal impact resulting from our work.
In terms of environmental impact, the short training times and consumer-grade GPU (RTX 3090/4090) required by our method makes it very economic in terms of power consumption and hardware requirements:
Overall, after downloading and setting up our code, running a single reconstruction, including data preprocessing with user inputs, takes under one hour of time.
When running on a slightly above average PSU with 750W at full capacity (which is usually not the case), given an average energy price of \$0.173 per kWh (April 2024 in the U.S.), we get a worst-case cost of \$0.13 per scene reconstruction.
Thus, in comparison to DynIBaR~\cite{li2023CVPR}, the current the state-of-the-art in terms of image quality, which requires 384 GPU hours on high-end A100 GPUs, our method has a significantly lower environmental footprint.
%
\end{document}